\pdfoutput=1

\documentclass[11pt]{article}

\usepackage[final]{acl}

\usepackage{times}
\usepackage{latexsym}

\usepackage[T1]{fontenc}

\usepackage[utf8]{inputenc}

\usepackage{microtype}

\usepackage{inconsolata}

\usepackage{graphicx}

\usepackage{xcolor}
\usepackage{tikz}

\usepackage{algorithm}
\usepackage{algpseudocode}
\usepackage{amsmath,amssymb,amsfonts}
\usepackage{graphicx}
\usepackage{bbold} 
\usepackage{booktabs}
\usepackage{multirow}
\usepackage{subcaption}

\newcommand{\redold}[1]{\textcolor{black}{ #1}}

\newcommand{\bluebold}[1]{\textcolor{blue}{\textbf{#1}}}
\newcommand{\purplebold}[1]{\textcolor{purple}{\textbf{#1}}}

\newcommand{\graycomment}[1]{\textcolor{lightgray}{\Comment{\parbox[t]{.5\linewidth}{#1}}}}

\title{MultiMatch: Multihead Consistency Regularization Matching for Semi-Supervised Text Classification\thanks{This paper has been accepted for publication in the Proceedings of the 2025 Conference on Empirical Methods in Natural Language Processing (EMNLP 2025).}}

\author{
 \textbf{Iustin Sirbu\textsuperscript{1,4}},
 \textbf{Robert-Adrian Popovici\textsuperscript{1}},\\
 \textbf{Cornelia Caragea\textsuperscript{2}},
 \textbf{Stefan Trausan-Matu\textsuperscript{1}},
 \textbf{Traian Rebedea\textsuperscript{1,3}}
\\
\\
 \textsuperscript{1}National University of Science and Technology POLITEHNICA Bucharest,\\
 \textsuperscript{2}University of Illinois Chicago,\hspace{0.1cm}
 \textsuperscript{3}NVIDIA,\hspace{0.1cm}
  \textsuperscript{4}Renius Technologies
\\
 \small{
  \href{mailto:iustin.sirbu@upb.ro}{\tt iustin.sirbu@upb.ro}
 }
}

\begin{document}

\maketitle
\begin{abstract}


We introduce \textbf{MultiMatch}, a novel semi-supervised learning (SSL) algorithm combining the paradigms of co-training and consistency regularization with pseudo-labeling. At its core, MultiMatch features a pseudo-label weighting module designed for selecting and filtering pseudo-labels based on head agreement and model confidence, and weighting them according to the perceived classification difficulty. 
This novel module enhances and unifies three existing techniques -- heads agreement from 
\textbf{Multi}head Co-training, self-adaptive thresholds from Free\textbf{Match}, and Average Pseudo-Margins from Margin\textbf{Match} -- resulting in a holistic approach that improves robustness and performance in SSL settings.
Experimental results on benchmark datasets highlight the superior performance of MultiMatch, i.e., MultiMatch achieves state-of-the-art results on 8 out of 10 setups from 5 natural language processing datasets and ranks first according to the Friedman test among 21 methods. Furthermore, MultiMatch demonstrates exceptional robustness in highly imbalanced settings, outperforming the second-best approach by 3.26\%, a critical advantage for real-world text classification tasks.
Our code is available on \href{https://github.com/iustinsirbu13/MultiMatch}{GitHub}.

\end{abstract}

\section{Introduction}
\vspace{-2mm}
Deep learning models typically require vast amounts of labeled data to achieve high performance across various tasks \cite{vaswani2017attention}. Moreover, the process of data annotation is both costly and time-intensive, often posing a significant bottleneck in real-world applications. To mitigate this challenge, semi-supervised learning methods (SSL) have emerged as a promising alternative. By leveraging a small set of labeled examples alongside a much larger unlabeled corpus, these approaches greatly reduce the time and resources needed for manual annotation while still maintaining competitive model performance~\cite{berthelot2019mixmatch}.

In recent years, semi-supervised learning has been dominated by two main paradigms. The first, pseudo-labeling with consistency regularization \cite{sohn2020fixmatch,zhang2022flexmatch}, enforces prediction invariance across different augmentations of the same sample, leveraging high-confidence predictions on weakly augmented data to generate pseudo-labels for strongly augmented counterparts. The second, co-training \cite{chen2021semisupervised,zou2023jointmatch}, mitigates error accumulation by training two networks in parallel, each providing pseudo-labels for the other.

Both paradigms have seen significant advancements. In pseudo-labeling with consistency regularization, research has primarily focused on improving pseudo-label filtering, with FreeMatch \cite{wang2023freematch} introducing class-wise self-adaptive thresholding and MarginMatch \cite{sosea2023marginmatch} incorporating predictions from past epochs to enhance reliability. In the co-training paradigm, Multihead Co-training \cite{chen2021semisupervised} reduced the need for multiple networks by employing a multi-headed architecture, while JointMatch \cite{zou2023jointmatch} highlighted the importance of samples where models produce different predictions. However, limited research has explored the integration of these paradigms to fully leverage their complementary strengths.

To bridge this gap, we introduce \textbf{MultiMatch}, a novel algorithm that incorporates a three-fold pseudo-label weighting module (PLWM).
This module integrates the voting mechanism from \textbf{Multi}head Co-training \cite{chen2021semisupervised} with a self-adaptive class-wise filtering strategy that leverages both historical and current predictions. Specifically, it incorporates the Average Pseudo-Margin (APM) introduced by Margin\textbf{Match} \cite{sosea2023marginmatch} to refine pseudo-label filtering based on historical predictions, while also utilizing the self-adaptive filtering mechanism of Free\textbf{Match} \cite{wang2023freematch}, which dynamically adjusts thresholds based on the model’s current predictions. 
\redold{Additionally, we define a taxonomy of “not useful”, “useful \& difficult”, and “useful \& easy” examples, 
and propose a weighting mechanism based on it.
}

We evaluate MultiMatch against 20 baselines across five text classification datasets from the unified SSL benchmark -- USB~\cite{wang2022usb}. To replicate the class imbalance challenges commonly encountered in real-world text classification tasks, we also construct highly imbalanced versions of the five datasets.
Our algorithm consistently outperforms all baselines, ranking first according to both the Friedman test and the mean error across all datasets, in both balanced and imbalanced settings.



To sum up, our main contributions are:
1) We introduce MultiMatch, a novel SSL algorithm that enhances and combines the strengths of different SSL paradigms; 2) We obtain state-of-the-art performance on five (out of five) datasets widely used for text classification; 3) We show the robustness of MultiMatch on skewed class distributions by obtaining state-of-the-art results in highly imbalanced settings; and 4) We provide a comprehensive ablation study and analysis of our approach.


\section{Related Work}
\label{sec:related_work}
\vspace{-1mm}
\paragraph{Semi-supervised Learning.}
Following the introduction of FixMatch \cite{sohn2020fixmatch}, a widely adopted paradigm in semi-supervised learning has been the combination of consistency regularization and pseudo-labeling. Consistency regularization \cite{sajjadi2016regularization} is based on the principle that the predictions of a model should remain invariant under realistic perturbations applied to input samples.
Pseudo-labeling \cite{mclachlan1975iterative} leverages the model's high-confidence predictions on unlabeled data, treating them as pseudo-labels that are 
incorporated into the training process.

Early approaches such as FixMatch \cite{sohn2020fixmatch} and UDA \cite{xie2020unsupervised_UDA} utilize a predefined threshold for the model's confidence to determine when its predictions can be converted into pseudo-labels. However, this method introduces two significant limitations. First, it does not account for the varying levels of difficulty across different classes, resulting in the model generating more pseudo-labels for the easier classes, thus creating a bias toward those classes \cite{zhang2022flexmatch, wang2023freematch}. Second, the model's confidence tends to be lower during the initial stages of training and increases as training progresses. As a result, this approach starts by generating few pseudo-labeled examples and then gradually produces high-confidence pseudo-labels for almost all samples later on, which can lead to error accumulation 
 and low generalization~\cite{arazo2020pseudo}. 

These limitations have been addressed by subsequent work in various ways. For instance, FlexMatch \cite{zhang2022flexmatch} and FreeMatch \cite{wang2023freematch} propose to adaptively adjust class-wise thresholds based on the estimated learning status of different classes.
MarginMatch \cite{sosea2023marginmatch} introduces the Average Pseudo-Margin (APM) score, which tracks the historical predictions for each sample and offers a more reliable confidence estimate. VerifyMatch \cite{park-caragea-2024-verifymatch} uses LLMs to assign initial pseudo-labels to unlabeled data, which are then verified with an SSL model for accuracy. 
Other works employ the co-training \cite{blum1998combining} framework. \citet{qiao2018deep} aim to alleviate error accumulation by using two distinct networks that are trained simultaneously, each one producing pseudo-labels for the other. Similarly, \citet{sadat-caragea-2024-co} employ two distinct networks that are trained simultaneously but instead the networks exchange sample importance weights (not pseudo-labels). 
JointMatch \cite{zou2023jointmatch} 
incorporates adaptive local thresholding similar to FreeMatch and introduces a weighting mechanism for disagreement examples, which are shown to be more informative. Multihead Co-training \cite{chen2021semisupervised} addresses the limitation of having to train multiple classifiers by proposing a multihead architecture where each head uses the agreement of the other ones as pseudo-label.


MultiMatch proposes to harness the strengths of both co-training and consistency regularization. While JointMatch has taken initial steps in this direction, the combination of a multihead architecture with a self-adaptive filtering mechanism based on historical predictions remains unexplored.

\vspace{-2mm}
\paragraph{Semi-supervised Text Classification.}
SSL has gained significant attention in text classification due to its ability to leverage large-scale unlabeled data alongside limited labeled examples using various approaches. 
For example, MixText \cite{chen2020mixtext} mitigates overfitting on limited data by integrating Mixup \cite{zhang2018mixup} to interpolate labeled and unlabeled examples in a hidden space. 
UDA \cite{xie2020unsupervised_UDA} uses strong data augmentation and consistency regularization to enforce prediction stability across perturbed inputs. 
AUM-ST \cite{sosea-caragea-2022-leveraging,gyawali-etal-2024-gunstance} refines self-training by filtering unreliable pseudo-labels using the Area Under the Margin \cite{pleiss2020identifying}, while SAT \cite{chen2022sat} enhances FixMatch \cite{sohn2020fixmatch} by re-ranking weak and strong augmentations based on input similarity. \citet{zou-etal-2023-decrisismb} propose an SSL approach that uses memory banks to handle imbalanced data. 
Additionally, co-training-based approaches have been explored. For example, \citet{hosseini2023semi} train two models in parallel -- one leveraging unsupervised domain adaptation and the other using SSL in the target domain -- allowing them to iteratively refine each other’s predictions. 



\section{MultiMatch}
\label{sec:our_approach}

\paragraph{Notation.} Let $\mathcal{X} = \{(x_b, y_b) ; b \in\{1, 2, ..., B\}\}$ be a batch of size $B$ of labeled examples and $\mathcal{U} = \{u_b; b \in \{1, 2, ..., \mu B\}\}$ be a batch of unlabeled examples, where $\mu$ is the ratio of labeled to unlabeled data in a batch. Let $\alpha(\cdot)$ and $\mathcal{A(\cdot)}$ be the weak and strong augmentation functions, respectively. The model's predictive distribution for a given input is denoted by $q = p_m(y|\cdot)$ and $\hat{q} = argmax(q)$ is the one-hot encoded pseudo-label. $\mathcal{H}(\cdot, \cdot)$ represents the cross-entropy function.

\subsection{Background}
\label{sec:background}

FixMatch \cite{sohn2020fixmatch} introduced the most popular recent direction in SSL which consists of combining consistency regularization \cite{sajjadi2016regularization} and pseudo-labeling \cite{mclachlan1975iterative}.


Specifically, for an unlabeled sample $u_b$, FixMatch creates two augmented versions using weak and strong augmentation functions, $\alpha(\cdot)$ and $\mathcal{A(\cdot)}$ respectively. Both $\alpha(u_b)$ and $\mathcal{A}(u_b)$ are passed through the model to obtain predictions $q_b = p_m(y|\alpha(u_b))$ and $Q_b = p_m(y|\mathcal{A}(u_b))$. The prediction on the weakly-augmented version is converted into a pseudo-label $\displaystyle \hat{q_{b}} = \arg\max_y(q_{b})$. FixMatch uses a predefined threshold $\tau$, such that only the high-confidence pseudo-labels that pass the filtering function $\mathbb{1}_{\tau}(q_{b}) = \mathbb{1}(\max(q_b) > \tau)$ are used in the unsupervised loss computation:
\vspace{-3mm}
\begin{equation}
    l_{u} = \frac{1}{\mu B} \sum_{b=1}^{\mu B} \mathbb{1}_{\tau}(q_{b}) H(\hat{q_{b}}, Q_{b})
\end{equation}

FlexMatch \cite{zhang2022flexmatch} and FreeMatch \cite{wang2023freematch} show that using a fixed confidence threshold $\tau$ is suboptimal, as it does not account for varying difficulty across classes and for the model's increase in confidence over time. They address these issues by introducing class-wise thresholds, adaptively scaling $\tau(c)$ depending on the learning status of each class. Specifically, FreeMatch computes a self-adaptive global threshold $\tau_t$ that reflects the confidence
of the model on the whole unlabeled dataset
and a self-adaptive local threshold $\tilde{p}_t(c)$ that reflects the confidence of the model for a specific class.
Then, the two thresholds are combined in order to obtain the final self-adaptive class thresholds $\tau_t(c)$ as the product between the normalized local thresholds and the global one. The computation of $\tau_t(c)$ is described formally in Appendix~\ref{sec:extended_background}.


These thresholds are used to replace the filtering function from FixMatch with:
\vspace{-1mm}
\begin{equation}
    \mathbb{1}_{Free}^{(t)} = \mathbb{1}(\max(q_b) > \tau_t(\hat{q_b}))
    \label{eq_freematch_filter}
\end{equation}

MarginMatch \cite{sosea2023marginmatch} builds on the previous approaches by showing that the current confidence of the model is not necessarily a good estimation for the quality of the pseudo-label. Instead, they propose that the historical beliefs should also be taken into account. To this end, they introduce the concept of pseudo-margin (PM) that is computed for each sample with respect to the pseudo-label $c$ at iteration $t$ as the difference between the logit $z_c$ corresponding to the assigned pseudo-label $c$ and the largest other logit: 

\vspace{-3mm}
\begin{equation}
    PM_c^{(t)}\redold{(u_b)} = z_c^{(t)} - \displaystyle \max_{i \ne c} z_i^{(t)}
    \label{eq_pm_formula}
\end{equation}

\noindent
Then, the Average Pseudo-Margin (APM) is computed using Exponential Moving Average 
using the smoothing parameter $\lambda_m$ to favor later iterations:

\vspace{-3mm}
\begin{equation}
\begin{aligned}[b]
    APM_c^{(t)}(\redold{u_b}) = PM_c^{(t)}(\redold{u_b}) \cdot \frac{\lambda_m}{1 + t} +\\ APM_c^{(t - 1)}(\redold{u_b}) \cdot (1 - \frac{\lambda_m}{1 + t})
    \end{aligned}
    \label{eq_apm_formula}
\end{equation}

\noindent
Finally, the APM confidence threshold $\gamma^{(t)}$ is computed for the current iteration using a a newly-introduced virtual class of erroneous examples and the filtering function becomes
$\mathbb{1}_{APM}^{(t)} = \mathbb{1}(APM_{\hat{q_b}}^{(t)}(u_b) > \gamma^{(t)})$.

\noindent

\begin{figure*}[!ht]
    \vspace{-3mm}
    \centering
    \scriptsize
    \captionsetup{font=small}
    \includegraphics[width=0.7\textwidth]{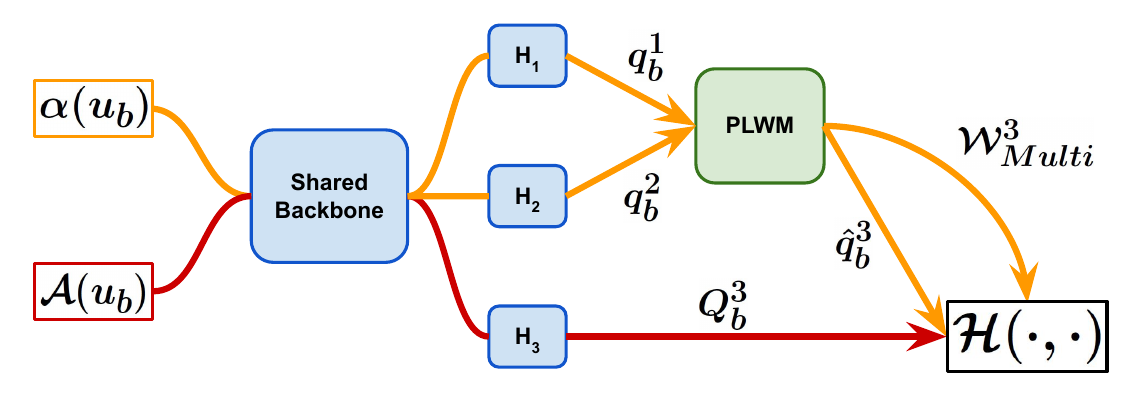}

    \caption{MultiMatch architecture \redold{with three heads}. Orange and red lines show the path of the weakly augmented and strongly augmented samples, respectively. The green PLWM module receives the $q_b^1$ and $q_b^2$ predictions on the weakly augmented samples from heads $H_1$ and $H_2$, then generates the pseudo-labels $\hat{q}_b^3$ and the corresponding weights $\mathcal{W}_{Multi}^3$. They are used together with the prediction $Q_b^3$ of head $H_3$ on the strongly augmented samples for computing the unsupervised loss $\mathcal{L}_u^3$ using the cross-entropy function. While the diagram shows the unlabeled loss computation for head $H_3$, similar computations are used for $H_1$ and $H_2$.}
    \label{fig:multimatch_arch}
    \vspace{-4mm}
\end{figure*}

\subsection{MultiMatch: Multihead Consistency Regularization Matching}
\label{sec:our_approach}

\textbf{MultiMatch} is our proposed semi-supervised learning algorithm that builds upon and extends elements from multiple SSL frameworks. The three primary sources of inspiration are \textbf{Multi}head Co-training \cite{chen2021semisupervised}, Margin\textbf{Match} \cite{sosea2023marginmatch}, and Free\textbf{Match} \cite{wang2023freematch}, which serve as representative approaches for the co-training and the consistency regularization with pseudo-labeling paradigms, respectively.

The pipeline for training a model with MultiMatch is shown in Figure \ref{fig:multimatch_arch}. 
Each batch of unlabeled examples is augmented using the weak and strong augmentation functions $\alpha(\cdot)$ and $\mathcal{A}(\cdot)$, respectively. The two augmented versions are forwarded through the model to make predictions. Note that, similar to Multihead Co-training, the model is multiheaded, which enables the co-training functionalities, while the shared backbone preserves the inference time of a single model. 
\redold{We use three heads in our presentation of the algorithm, both for simplicity and because \citet{chen2021semisupervised} identified this to be the optimal choice.}

As multiple predictions are obtained, pseudo-label selection is less straightforward than in classical pseudo-labeling, unless disagreement samples are discarded -- a strategy we consider suboptimal. To address this, we introduce a pseudo-label weighting module (PLWM), which serves three key functions: selecting pseudo-labels, filtering out unreliable pseudo-labels and weighting samples according to their estimated difficulty. These functions are achieved through our three-fold approach, which integrates and refines head agreements, self-adaptive thresholding, and APM scores.




The pseudocode of MultiMatch can be viewed in Algorithm \ref{multimatch_algorithm}. More details about the algorithm are provided in the remainder of this section.

\begin{algorithm*}
\small
\begin{algorithmic}[1]
\Require Labeled batch $\mathcal{X} = \{(x_b, y_b) ; b \in(1, 2, ..., B)\}$, unlabeled batch $\mathcal{U} = \{u_b; b = (1, 2, ..., \mu B)\}$, where $B$ is the batch size of labeled data, $\mu$ is the ratio of unlabeled to labeled data, unsupervised loss weight $w_u$, disagreement weight $\delta$, FreeMatch EMA decay $\lambda_f$, MarginMatch EMA decay $\lambda_m$, weak and strong augmentations $\alpha(\cdot)$, $\mathcal{A}(\cdot)$.

\vspace{2mm}
\For{ $h$ in $1$ to $3$} \graycomment{For each classification head.}
    \vspace{2mm}
    \State $i, j = \{1, 2, 3\} \setminus \{h\}$ \graycomment{The other 2 heads are used for generating pseudo-labels.}

    \vspace{2mm}
    \State $\mathcal{L}_s^h = \frac{1}{B}\sum_{b=1}^B\mathcal{H}(p_b, p_m^h(y | \alpha(x_b)))$ \graycomment{Compute the supervised loss for head $h$ using the labeled data.}




    \vspace{2mm}
    \State Compute $\mathbb{1}_{Free}^{(t)}(i)$ and $\mathbb{1}_{Free}^{(t)}(j)$ using \redold{Eq. \ref{eq_freematch_filter}} for the remaining heads $i$ and $j$.



    \vspace{2mm}
    \State Compute $APM_{k, c}^{(t)}(u_b)$ for all classes $c \in \{1,...,C\}$, heads $k \in \{i, j\} $ and all samples $u_b \in \mathcal{U}$ using Eq. \ref{eq_pm_formula} and \ref{eq_apm_formula}.



    \vspace{2mm}
    \State Select $\hat{q}_b^h$ and compute $W_{Multi}^h$ using Eq. \ref{eq_weights_multi}. \graycomment{Use PLWM to select, filter, and weight pseudo-labels.}

    \vspace{2mm}
    \State $\mathcal{L}_{u,t}^h = \frac{1}{\mu B} \sum_{b=1}^{\mu B} W_{Multi}^h    \cdot    H(\hat{q}_{b}^h, Q_{b}^h)$ \graycomment{Compute the unsupervised loss for head $h$ using the generated pseudo-labels and weights.}

\EndFor

\vspace{2mm}
\State Compute $\gamma_{k, c}^{(t)}$, $k \in \{1,2,3\}$, using Eq. \ref{eq_gamma_apm}. \graycomment{Update our self-adaptive APM thresholds.}

\vspace{2mm}

\State $\mathcal{L}_s = \sum_{k\in\{1,2,3\}}\mathcal{L}_s^k$ \hspace{2mm}and\hspace{2mm} $\mathcal{L}_u = \sum_{k\in\{1,2,3\}}\mathcal{L}_u^k$
\vspace{2mm}
\State $\mathcal{L} = \mathcal{L}_s + w_u \cdot\mathcal{L}_u$ \graycomment{Compute the total loss.}

\end{algorithmic}
\caption{MultiMatch}
\label{multimatch_algorithm}
\end{algorithm*}

\subsection{Pseudo-Label Weighting Module (PLWM)}

In the co-training framework, the training samples are usually part of two categories: the ones where the models agree on the prediction and the disagreement samples. However, different methods treat them in different ways. On one hand, some methods consider the agreement samples to be irrelevant, as they do not convey new information, so they are removed from the loss computation~\cite{coteachingplus} to prevent the convergence of the models. These methods only use the disagreement samples for training. On the other hand, the multihead methods consider the agreement samples to provide high-confidence pseudo-labels, so they use them and remove the disagreement samples from the loss computation \cite{chen2021semisupervised}. Finally, JointMatch \cite{zou2023jointmatch} tries to balance the two sides by using the agreement only as a weighting mechanism, while the filtering itself is done by a distinct confidence-based heuristic. 

We introduce a new mechanism of using the agreement of the heads. By combining it with the APM score, we split the generated pseudo-labels into 3 categories: \textit{not useful}, \textit{useful \& difficult}, and \textit{useful \& easy}.

If the heads of the model used to generate the pseudo-label reach agreement \redold{($\hat{q}_{b}^i = \hat{q}_{b}^j$)} and both have high confidence in terms of APM score, the sample is considered \textit{useful \& easy}, so it is included in the loss computation with weight $1$\redold{, using the agreed pseudo-label}.
If one of the heads has high confidence and the other one has low confidence, in terms of APM score, we consider that sample to be \textit{useful \& difficult}, as one head is not confident in the prediction. 
\redold{In this case, the selected pseudo-label is the one generated by the confident head. We include it in the loss computation with a larger weight $w_d$.}
If both heads have low confidence or if both heads have high confidence but \redold{disagree on the label}, we consider that sample to be \textit{not useful} for the current iteration and assign it weight $0$. \redold{This is equivalent to filtering out the sample.} To define high confidence in terms of the APM score, we introduce a novel method for computing self-adaptive APM thresholds, which is detailed in Section \ref{subsec:self-adaptive-apm-thresholds}.
Finally, we employ an additional self-adaptive filtering based on the current confidence of each head.
The computation of the weights $W_{Multi}^h$ for the head $h$ and the unlabeled batch $u_b$ is described formally in Eq. \ref{eq_weights_multi}.



\vspace{-5mm}
\begin{equation}
\begin{aligned}[b]
    W_{Multi}^h =  [1 \cdot (\mathbb{1}_{Multi}^i \land \mathbb{1}_{Multi}^j \land \mathbb{1}_{Agree}^h) + \\ + w_d \cdot (\mathbb{1}_{Multi}^i \oplus \mathbb{1}_{Multi}^j)] \cdot \mathbb{1}_{FreeMulti}^{h}
\end{aligned}
\label{eq_weights_multi}
\end{equation}

\noindent
where $\mathbb{1}_{Multi}^k = \mathbb{1}(APM_{\hat{q}_b^k}^{(t)}(u_b) > \gamma_{k, c}^{(t-1)})$, $k \in \{i, j\}$ marks the high confidence of head $k$, $\mathbb{1}_{Agree}^h = \redold{[ \hat{q}_b^i = \hat{q}_{b}^j]}, \{i, j\} = \{1, 2, 3\} \setminus \{ h \}$ is the agreement of the remaining heads $i$ and $j$, and $\mathbb{1}_{FreeMulti}^{h} = \mathbb{1}_{Free}^{i} \vee \mathbb{1}_{Free}^{j}$ is the self-adaptive threshold filtering based on the current confidence, adapted to multiple heads.

\subsection{Self-adaptive APM Thresholds}
\label{subsec:self-adaptive-apm-thresholds}

The use of Average-Pseudo Margins~\cite{sosea2023marginmatch} has shown great potential in addressing the common issues that come with relying on the current model's confidence for evaluating the quality of pseudo-labels. However, the thresholding approach used by MarginMatch still has its limits, as the adaptive threshold $\gamma^{(t)}$ is the same for all classes. 
We address this by using class-wise thresholds that take into account the different learning difficulty of each class.

Furthermore, we take advantage of our multihead architecture and instead of introducing an additional virtual class, we use the subset of examples where the heads reached agreement as the subset used for the threshold estimation. 

By starting from a subset assumed to be correctly pseudo-labeled and choosing the $f=5^{th}$ percentile, we obtain a threshold estimate 
that favors correctly pseudo-labeled samples,
minimizing the number of incorrect pseudo-labels that pass the threshold. On the other hand, MarginMatch starts from an incorrectly classified subset and uses the $95^{th}$ percentile value, thus potentially choosing a threshold that allows for more incorrect pseudo-labels to pass, leading to error accumulation for the model.
Finally, adhering to our principle of minimizing the noisy pseudo-labels, we set a lower bound $\gamma_{{min}} = 0$ for the computed threshold.

Eq. \ref{eq_gamma_apm} shows the formula for choosing the threshold $\gamma_{h,c}^{(t)}$ for head $h$ and class $c$ at iteration $t$ by using the agreement of the other heads $k \neq h$:
\begin{equation}
\begin{aligned}[b]
    \gamma_{\redold{h},c}^{(t)} = max({\gamma_{min}}, \text{percentile}_{f}( \hspace{20mm} \\
    \displaystyle \cup_{k \ne h}
    \{ APM_{k,c}^{(t)} (u_b) \mid 
    \mathbb{1}_{Agree}^{h,c} \})) 
    \label{eq_gamma_apm}
\end{aligned}
\end{equation}


\noindent
where $\mathbb{1}_{Agree}^{h, c} = [ \redold{\hat{q}_{b}^i = \hat{q}_{b}^j} = c], \{i, j\} = \{1, 2, 3\} \setminus \{ h \}$ marks if the remaining two heads have agreed on the pseudo-label $c$.

\begin{table*}[!htbp]
\scriptsize
\centering 
\captionsetup{font=small}
\tabcolsep=0.1cm

\begin{tabular}{l|rr|rr|rr|rr|rr|r|c|c}
\toprule

Dataset & \multicolumn{2}{c|}{IMDB} & \multicolumn{2}{c|}{AG News} & \multicolumn{2}{c|}{Amazon Review} & \multicolumn{2}{c|}{Yahoo! Answer} & \multicolumn{2}{c|}{Yelp Review}  & Mean & Fried. & Final \\

\# Label & 20 & 100 & 40 & 200 & 250 & 1000 & 500 & 2000 & 250 & 1000 & error & rank & rank \\

\midrule

Supervised (Full) & 5.69\scalebox{.6}{±0.15} & 5.72\scalebox{.6}{±0.13} & 5.78\scalebox{.6}{±0.07} & 5.73\scalebox{.6}{±0.11} & 36.40\scalebox{.6}{±0.05} & 36.40\scalebox{.6}{±0.05} & 24.87\scalebox{.6}{±0.04} & 24.84\scalebox{.6}{±0.04} & 32.04\scalebox{.6}{±0.03} & 32.04\scalebox{.6}{±0.03} & 20.95 & - & -\\
Supervised (Small) & 20.31\scalebox{.6}{±2.79} & 14.02\scalebox{.6}{±1.22} & 15.06\scalebox{.6}{±1.08} & 14.25\scalebox{.6}{±0.97} & 52.31\scalebox{.6}{±1.28} & 47.53\scalebox{.6}{±0.69} & 37.43\scalebox{.6}{±0.29} & 33.26\scalebox{.6}{±0.10} & 51.22\scalebox{.6}{±0.98} & 46.71\scalebox{.6}{±0.37} & 33.21 & - & -\\
\midrule
AdaMatch & 8.09\scalebox{.6}{±0.99} & 7.11\scalebox{.6}{±0.20} & 11.73\scalebox{.6}{±0.17} & 11.22\scalebox{.6}{±0.95} & 46.72\scalebox{.6}{±0.72} & 42.27\scalebox{.6}{±0.25} & 32.75\scalebox{.6}{±0.35} & 30.44\scalebox{.6}{±0.31} & 45.40\scalebox{.6}{±0.96} & 40.16\scalebox{.6}{±0.49} & 27.59 & \hphantom{0}6.5 & \hphantom{0}5 \\
FixMatch & 7.72\scalebox{.6}{±0.33} & 7.33\scalebox{.6}{±0.13} & 30.17\scalebox{.6}{±1.87} & 11.71\scalebox{.6}{±1.95} & 47.61\scalebox{.6}{±0.83} & 43.05\scalebox{.6}{±0.54} & 33.03\scalebox{.6}{±0.49} & 30.51\scalebox{.6}{±0.53} & 46.52\scalebox{.6}{±0.94} & 40.65\scalebox{.6}{±0.46} & 29.83 & 10.2 & 11 \\
FlexMatch & 7.82\scalebox{.6}{±0.77} & 7.41\scalebox{.6}{±0.38} & 16.38\scalebox{.6}{±3.94} & 12.08\scalebox{.6}{±0.73} & 45.73\scalebox{.6}{±1.60} & 42.25\scalebox{.6}{±0.33} & 35.61\scalebox{.6}{±1.08} & 31.13\scalebox{.6}{±0.18} & 43.35\scalebox{.6}{±0.69} & 40.51\scalebox{.6}{±0.34} & 28.23 & \hphantom{0}9.4 & \hphantom{0}9 \\
Dash & 8.34\scalebox{.6}{±0.86} & 7.55\scalebox{.6}{±0.35} & 31.67\scalebox{.6}{±13.1} & 13.76\scalebox{.6}{±1.67} & 47.10\scalebox{.6}{±0.74} & 43.09\scalebox{.6}{±0.60} & 35.26\scalebox{.6}{±0.33} & 31.19\scalebox{.6}{±0.29} & 45.24\scalebox{.6}{±2.02} & 40.14\scalebox{.6}{±0.79} & 30.33 & 11.7 & 13 \\
CRMatch & 8.96\scalebox{.6}{±0.88} & 7.16\scalebox{.6}{±0.09} & 12.28\scalebox{.6}{±1.43} & 11.08\scalebox{.6}{±1.24} & 45.49\scalebox{.6}{±0.98} & 43.07\scalebox{.6}{±0.50} & 32.51\scalebox{.6}{±0.40} & 29.98\scalebox{.6}{±0.07} & 45.71\scalebox{.6}{±0.63} & 40.62\scalebox{.6}{±0.28} & 27.69 & \hphantom{0}7.6 & \hphantom{0}6 \\
CoMatch & 7.44\scalebox{.6}{±0.30} & 7.72\scalebox{.6}{±1.14} & 11.95\scalebox{.6}{±0.76} & 10.75\scalebox{.6}{±0.35} & 48.76\scalebox{.6}{±0.90} & 43.36\scalebox{.6}{±0.21} & 33.48\scalebox{.6}{±0.51} & 30.25\scalebox{.6}{±0.35} & 45.40\scalebox{.6}{±1.12} & 40.27\scalebox{.6}{±0.51} & 27.94 & \hphantom{0}8.1 & \hphantom{0}8 \\
SimMatch & 7.93\scalebox{.6}{±0.55} & 7.08\scalebox{.6}{±0.33} & 14.26\scalebox{.6}{±1.51} & 12.45\scalebox{.6}{±1.37} & 45.91\scalebox{.6}{±0.95} & 42.21\scalebox{.6}{±0.30} & 33.06\scalebox{.6}{±0.20} & 30.16\scalebox{.6}{±0.21} & 46.12\scalebox{.6}{±0.48} & 40.26\scalebox{.6}{±0.62} & 27.94 & \hphantom{0}7.6 & \hphantom{0}6 \\
FreeMatch & 8.94\scalebox{.6}{±0.21} & 7.95\scalebox{.6}{±0.45} & 12.98\scalebox{.6}{±0.58} & 11.73\scalebox{.6}{±0.63} & 46.41\scalebox{.6}{±0.60} & 42.64\scalebox{.6}{±0.06} & 32.77\scalebox{.6}{±0.26} & 30.32\scalebox{.6}{±0.18} & 47.95\scalebox{.6}{±1.45} & 40.37\scalebox{.6}{±1.00} & 28.21 & \hphantom{0}9.7 & 10 \\
SoftMatch & 7.76\scalebox{.6}{±0.58} & 7.97\scalebox{.6}{±0.72} & 11.90\scalebox{.6}{±0.27} & 11.72\scalebox{.6}{±1.58} & 45.29\scalebox{.6}{±0.95} & 42.21\scalebox{.6}{±0.20} & 33.07\scalebox{.6}{±0.31} & 30.44\scalebox{.6}{±0.62} & 44.09\scalebox{.6}{±0.50} & 39.76\scalebox{.6}{±0.13} & 27.42 & \hphantom{0}6.4 & \hphantom{0}4 \\
\midrule
Multihead Co-training & 8.70\scalebox{.6}{±0.88} & 7.46\scalebox{.6}{±0.68} & 22.72\scalebox{.6}{±6.09} & 13.48\scalebox{.6}{±1.51} & 46.22\scalebox{.6}{±0.70} & 43.07\scalebox{.6}{±0.79} & 35.17\scalebox{.6}{±1.58} & 30.81\scalebox{.6}{±0.28} & 46.46\scalebox{.6}{±1.28} & 40.79\scalebox{.6}{±0.49} & 29.49 & 12.2 & 14 \\
MarginMatch & 7.19\scalebox{.6}{±0.39} & 6.99\scalebox{.6}{±0.19} & \bluebold{10.65\scalebox{.6}{±0.19}} & 11.03\scalebox{.6}{±0.99} & 44.81\scalebox{.6}{±1.23} & 42.14\scalebox{.6}{±0.67} & 32.08\scalebox{.6}{±0.70} & 29.55\scalebox{.6}{±0.15} & 42.93\scalebox{.6}{±1.48} & 39.13\scalebox{.6}{±0.34} & 26.65 & \hphantom{0}2.3 & \hphantom{0}2 \\

SequenceMatch & 12.99\scalebox{.6}{±7.64} & 8.34\scalebox{.6}{±0.57} & 16.17\scalebox{.6}{±2.53} & 12.56\scalebox{.6}{±0.87} & 47.97\scalebox{.6}{±2.39} & 42.58\scalebox{.6}{±0.21} & 34.92\scalebox{.6}{±0.78} & 30.28\scalebox{.6}{±0.28} & 45.60\scalebox{.6}{±0.84} & 40.44\scalebox{.6}{±0.51} & 29.19 & 11.3 & 12 \\
CGMatch & 7.07\scalebox{.6}{±0.36} & \bluebold{6.79\scalebox{.6}{±0.10}} & 11.95\scalebox{.6}{±0.63} & 11.29\scalebox{.6}{±0.47} & 44.77\scalebox{.6}{±1.39} & 42.61\scalebox{.6}{±0.47} & 32.15\scalebox{.6}{±0.58} & 29.85\scalebox{.6}{±0.16} & 44.34\scalebox{.6}{±0.26} & 40.14\scalebox{.6}{±0.07} & 27.10 & \hphantom{0}3.9 & \hphantom{0}3 \\

\midrule
MultiMatch & \bluebold{6.89\scalebox{.6}{±0.07}} & 6.98\scalebox{.6}{±0.27} & 11.14\scalebox{.6}{±0.96} & \bluebold{10.59\scalebox{.6}{±0.66}} & \bluebold{44.43\scalebox{.6}{±0.98}} & \bluebold{42.09\scalebox{.6}{±0.28}} & \bluebold{30.90\scalebox{.6}{±0.70}} & \bluebold{29.39\scalebox{.6}{±0.39}} & \bluebold{42.16\scalebox{.6}{±0.79}} & \bluebold{39.08\scalebox{.6}{±0.55}} & \bluebold{26.37} & \hphantom{0}\bluebold{1.2} & \hphantom{0}\bluebold{1} \\

\bottomrule
\end{tabular}
\caption{Test error rates on IMDB, AG News, Amazon Review, Yahoo! Answer, and Yelp Review datasets using two setups with different sizes for the labeled set. The best result for each setup is highlighted in \bluebold{blue}. The full table is shown in Appendix \ref{sec:appendix}.}
\label{tab:full-results}
\vspace{-4mm}
\end{table*}

\section{Experiments}
\label{sec:experiments}

We assess the performance of our MultiMatch algorithm across a diverse set of SSL benchmark datasets. Specifically, we utilize the USB benchmark \cite{wang2022usb}, which offers a standardized evaluation framework for semi-supervised classification across vision, language, and audio tasks.
We perform experiments with various numbers of labeled examples on the natural language processing (NLP) datasets included in USB: IMDB \cite{maas2011learning_imdb}, AG News \cite{zhang2015character_agnews}, Amazon Review \cite{mcauley2013hidden_amazon}, Yahoo! Answer \cite{chang2008importance_yahoo} and Yelp Review \cite{asghar2016yelp}. 
For a straightforward and fair comparison with the 16 SSL algorithms already included in USB, we maintain their original experimental setup without modifications. This ensures consistency across evaluations and includes using the uncased version of BERT-Base \cite{devlin2018bert} for the shared backbone, preserving the train-validation-test splits, the number of labeled and unlabeled examples per class, and all training hyperparameters. 
We provide dataset statistics and split information in Appendix~\ref{sec:appendix_dataset}, and more details on the experimental setup in Appendix \ref{sec:hyperparams}.

For the supervised baselines and the SSL baselines implemented in USB, we report the results from the USB GitHub page\footnote{\scriptsize{\href{https://github.com/microsoft/Semi-supervised-learning/blob/main/results/usb_nlp.csv}{https://github.com/microsoft/Semi-supervised-learning}}}, which provide an improvement over those originally reported by \citet{wang2022usb} by using a larger batch size. For the other 4 baselines (Multihead Co-training, MarginMatch, SequenceMatch, CGMatch) and for MultiMatch we add the implementations to the USB framework and report the average test error of 3 runs per experiment (using the same seeds). In addition to the mean error rate across tasks, we compute the Friedman rank of each algorithm as the mean rank over all tasks $\text{rank}_F = \frac{1}{m}\sum_{i=1}^m\text{rank}_i$, where $m=10$ is the number of evaluation setups and $\text{rank}_i$ is the rank of the algorithm in the $i$-th setup.

\paragraph{Imbalanced settings.} 
While the research on semi-supervised learning in NLP widely uses balanced setups for evaluation, the class distributions of many real-world text classification datasets are imbalanced \cite{gui2024survey_cissl}.
Moreover, while it is well known that classifiers trained on imbalanced data have a bias toward the majority classes, this issue is even more problematic for SSL algorithms where the use of biased pseudo-labels increases the bias over time \cite{kim2020distribution}.

\begin{table*}[t]
\scriptsize
\centering 
\captionsetup{font=small}
\tabcolsep=0.1cm

\begin{tabular}{l|cc|cc|cc|cc|cc|c|c|c}
\toprule

Dataset & \multicolumn{2}{c|}{IMDB} & \multicolumn{2}{c|}{AG News} & \multicolumn{2}{c|}{Amazon Review} & \multicolumn{2}{c|}{Yahoo! Answer} & \multicolumn{2}{c|}{Yelp Review}  & Mean & Fried. & Final \\

Imbalance & 100 & -100 & 100 & -100 & 100 & -100 & 100 & -100 & 100 & -100 & error & rank & rank \\

\midrule

FixMatch & 49.95\scalebox{.6}{±0.06} & 49.92\scalebox{.6}{±0.14} & 31.95\scalebox{.6}{±5.30} & 36.09\scalebox{.6}{±6.50} & 64.69\scalebox{.6}{±0.92} & 61.93\scalebox{.6}{±4.41} & 51.64\scalebox{.6}{±0.75} & 52.78\scalebox{.6}{±1.45} & 65.57\scalebox{.6}{±3.19} & 57.43\scalebox{.6}{±6.22} & 52.20 & 8.4 & 10 \\

\hspace{5mm} + ABC & 49.69\scalebox{.6}{±0.43} & 45.90\scalebox{.6}{±7.07} & 38.08\scalebox{.6}{±8.80} & 21.21\scalebox{.6}{±5.72} & 62.02\scalebox{.6}{±1.03} & 56.27\scalebox{.6}{±1.16} & 61.15\scalebox{.6}{±3.12} & 53.79\scalebox{.6}{±9.31} & 62.63\scalebox{.6}{±0.89} & 58.37\scalebox{.6}{±6.97} & 50.91 & 6.3 & 7 \\
\midrule

FreeMatch & 49.95\scalebox{.6}{±0.03} & 42.75\scalebox{.6}{±12.56} & 26.55\scalebox{.6}{±4.87} & \bluebold{24.58\scalebox{.6}{±3.71}} & 62.26\scalebox{.6}{±0.48} & 59.22\scalebox{.6}{±1.94} & 44.93\scalebox{.6}{±1.00} & \bluebold{40.96\scalebox{.6}{±0.43}} & 63.92\scalebox{.6}{±3.98} & 59.13\scalebox{.6}{±1.08} & 47.43 & 6.0 & 6\\
\hspace{5mm} + ABC & 49.97\scalebox{.6}{±0.04} & 30.62\scalebox{.6}{±18.18} & 23.76\scalebox{.6}{±5.45} & 24.01\scalebox{.6}{±1.66} & 58.13\scalebox{.6}{±0.93} & \purplebold{54.86\scalebox{.6}{±0.34}} & 49.60\scalebox{.6}{±7.68} & 44.04\scalebox{.6}{±4.18} & 61.58\scalebox{.6}{±2.69} & 54.02\scalebox{.6}{±3.44} & 45.06 & 3.9 & 4 \\
\midrule

Multihead Co-training & 49.91\scalebox{.6}{±0.12} & 50.00\scalebox{.6}{±0.00} & 30.88\scalebox{.6}{±2.81} & 34.43\scalebox{.6}{±4.51} & 62.90\scalebox{.6}{±2.24} & \bluebold{58.05\scalebox{.6}{±4.53}} & 52.05\scalebox{.6}{±1.28} & 51.39\scalebox{.6}{±1.51} & 64.32\scalebox{.6}{±3.06} & 58.62\scalebox{.6}{±3.34} & 51.25 & 7.9 & 9 \\
\hspace{5mm} + ABC & 49.76\scalebox{.6}{±0.24} & 50.00\scalebox{.6}{±0.00} & 25.29\scalebox{.6}{±1.41} & 26.18\scalebox{.6}{±1.20} & \purplebold{57.99\scalebox{.6}{±3.54}} & 57.07\scalebox{.6}{±1.56} & 46.92\scalebox{.6}{±2.81} & 49.00\scalebox{.6}{±6.97} & 63.48\scalebox{.6}{±3.90} & 55.38\scalebox{.6}{±1.50} & 48.11 & 5.1 & 5 \\

\midrule

MarginMatch & 49.60\scalebox{.6}{±0.62} & 49.99\scalebox{.6}{±0.01} & 26.46\scalebox{.6}{±0.99} & 33.85\scalebox{.6}{±2.93} & 64.74\scalebox{.6}{±0.39} & 66.45\scalebox{.6}{±0.81} & 50.59\scalebox{.6}{±0.86} & 53.96\scalebox{.6}{±0.81} & 63.33\scalebox{.6}{±0.58} & 63.70\scalebox{.6}{±2.41} & 52.27 & 7.8 & 8 \\

\hspace{5mm} + ABC & \purplebold{45.04\scalebox{.6}{±4.14}} & 42.55\scalebox{.6}{±12.85} & \purplebold{23.74\scalebox{.6}{±2.77}} & \purplebold{17.88\scalebox{.6}{±0.75}} & 59.98\scalebox{.6}{±1.45} & 55.71\scalebox{.6}{±3.13} & 48.48\scalebox{.6}{±9.07} & 52.75\scalebox{.6}{±2.71} & 61.43\scalebox{.6}{±3.15} & 55.42\scalebox{.6}{±0.33} & 46.30 & 3.2 & 2 \\

\midrule

MultiMatch & \bluebold{49.17\scalebox{.6}{±0.88}} & \bluebold{26.05\scalebox{.6}{±6.28}} & \bluebold{21.11\scalebox{.6}{±0.49}} & 25.36\scalebox{.6}{±2.60} & \bluebold{61.03\scalebox{.6}{±1.64}} & 59.66\scalebox{.6}{±2.58} & \bluebold{41.01\scalebox{.6}{±0.79}} & 41.46\scalebox{.6}{±4.34} & \bluebold{60.14\scalebox{.6}{±3.41}} & \bluebold{56.71\scalebox{.6}{±1.87}} & \bluebold{44.17} & \bluebold{3.3} & \bluebold{3} \\

\hspace{5mm} + ABC & 48.81\scalebox{.6}{±1.98} & \purplebold{17.21\scalebox{.6}{±3.78}} & 29.36\scalebox{.6}{±11.90} & 19.65\scalebox{.6}{±4.39} & 62.13\scalebox{.6}{±1.46} & 57.12\scalebox{.6}{±0.88} & \purplebold{41.05\scalebox{.6}{±1.34}} & \purplebold{42.72\scalebox{.6}{±2.57}} & \purplebold{60.71\scalebox{.6}{±2.68}} & \purplebold{53.87\scalebox{.6}{±3.03}} & \purplebold{43.26} & \purplebold{3.1} & \purplebold{1} \\

\bottomrule
\end{tabular}
\caption{Test error rates in imbalanced setups. The best results \bluebold{without}/\purplebold{with} ABC are highlighted in \bluebold{blue}/\purplebold{purple}. Imbalance 100: similar distributions for labeled and unlabeled data; Imbalance -100: reversed long-tail distribution for the unlabeled set. }
\label{tab:imbalanced-results}
\vspace{-4mm}
\end{table*}

In order to assess the robustness of MultiMatch when data is imbalanced, we draw inspiration from the setup used by ABC in Computer Vision~\cite{lee2021abc} to create class-imbalanced versions of the text datasets from USB. As we focus on high-imbalance scenarios, we consider a long-tailed imbalance, where the number of data points decreases exponentially from the largest to the smallest class. Formally, the number of samples from class $c$ is $N_c = N_1 \cdot \gamma^{-\frac{c-1}{C-1}}$, where $C$ is the number of classes, $c\in \{1,...,C\}$, and $\gamma = \frac{N_1}{N_C}$ is the imbalance factor between the largest and the smallest class.
For each dataset, we consider two highly imbalanced setups: first, we consider a long-tail imbalance that is similar in the labeled and unlabeled datasets. In this case, we set the imbalance factor $\gamma = 100$, the size of the largest class $N_1 = 1000$ and of the smallest class $N_C = 10$ and the unlabeled set to be 10 times larger than the labeled one. For the second setup, we keep the same imbalance for the labeled dataset, but we use an unlabeled imbalance factor of $\gamma = -100$, which represents a long-tail distribution where the most represented classes in the labeled dataset are the least represented ones in the unlabeled set. This represents an extreme case where the unlabeled corpus is collected separately from the labeled dataset and the class distributions might be entirely different.


Since SSL algorithms exhibit significantly different performance in imbalanced settings compared to balanced ones, and the performance gap between algorithms is larger in imbalanced setups, we integrate a class-imbalance semi-supervised learning (CISSL) method, ABC~\cite{lee2021abc}, to assess the potential of bridging this gap. ABC accounts for class imbalance by attaching an additional balanced classifier, without discarding any data. Given its seamless integration with existing SSL algorithms, we conduct additional experiments by enhancing each tested SSL algorithm with ABC in imbalanced settings. Further details on CISSL and ABC are provided in the Appendix \ref{sec:appendix:abc}.

\section{Results}
\label{sec_results}

The comparison between MultiMatch and other SSL methods on the five text classification datasets is presented in Table~\ref{tab:full-results}. Our results show that MultiMatch consistently outperforms all the baselines, achieving top performance in 8 out of 10 setups and ranking first according to the Friedman test among 21 SSL algorithms.
While the overall mean error also decreases by an absolute value of 0.28\% compared to the second best approach, we notice that the improvement is higher when the number of labeled examples is lower. Specifically, the improvement over MarginMatch is on average 0.43\% in the lower-resource setting, compared to 0.14\% when more labeled data is provided.

By analyzing the baselines that inspired our approach, we observe that MarginMatch consistently outperforms all methods except MultiMatch, reinforcing the importance of leveraging historical evidence for pseudo-labels filtering. In contrast, Multihead Co-training and FreeMatch have a lower rank, $14$ and $10$, respectively, indicating that the voting mechanism and the self-adaptive thresholds, while beneficial, are not among the most impactful techniques when used in isolation. However, when refined and incorporated into our unique three-fold pseudo-label weighting module (PLWM), their combined effect leads to substantial performance gains across all evaluation setups.

\paragraph{Imbalanced settings.}

We evaluate MultiMatch and representatives of each incorporated technique in newly created highly imbalanced setups of the same datasets from USB. 
Specifically, as shown in Table~\ref{tab:imbalanced-results}, we compare MultiMatch with MarginMatch, representing the Average Pseudo-Margins approach; Multihead Co-training, embodying the voting mechanism from co-training; FreeMatch, representing the self-adaptive class thresholds technique; and FixMatch, serving as a baseline for pseudo-labeling with consistency regularization.

\begin{table*}[!htbp]
\scriptsize
\centering 
\captionsetup{font=small}
\tabcolsep=0.1cm

\begin{tabular}{l|cc|cc|cc|cc|cc|c|c|c}
\toprule

Dataset & \multicolumn{2}{c|}{IMDB} & \multicolumn{2}{c|}{AG News} & \multicolumn{2}{c|}{Amazon Review} & \multicolumn{2}{c|}{Yahoo! Answer} & \multicolumn{2}{c|}{Yelp Review} & Mean & Fried. & Final \\

\# Label & 20 & 100 & 40 & 200 & 250 & 1000 & 500 & 2000 & 250 & 1000 & error & rank & rank \\

\midrule
MultiMatch & 6.89\scalebox{.6}{±0.07} & \bluebold{6.98\scalebox{.6}{±0.27}} & 11.14\scalebox{.6}{±0.96} & 10.59\scalebox{.6}{±0.66} & 44.43\scalebox{.6}{±0.98} & 42.09\scalebox{.6}{±0.28} & \bluebold{30.90\scalebox{.6}{±0.70}} & \bluebold{29.39\scalebox{.6}{±0.39}} & \bluebold{42.16\scalebox{.6}{±0.79}} & \bluebold{39.08\scalebox{.6}{±0.55}} & \bluebold{26.37} & \bluebold{1.8} & \bluebold{1} \\

\hspace{3mm} --- $w_d$ & 7.08\scalebox{.6}{±0.32} & 7.01\scalebox{.6}{±0.31} & \bluebold{10.93\scalebox{.6}{±0.43}} & 10.73\scalebox{.6}{±0.62} & 44.47\scalebox{.6}{±1.11} & 42.17\scalebox{.6}{±0.18} & 31.24\scalebox{.6}{±0.47} & 29.64\scalebox{.6}{±0.51} & 44.06\scalebox{.6}{±1.28} & 39.41\scalebox{.6}{±0.46} & 26.67 & 3.7 & 4 \\

\hspace{3mm} --- $w_d = 0$ & 7.19\scalebox{.6}{±0.29} & 7.20\scalebox{.6}{±0.17} & 11.80\scalebox{.6}{±0.88} & 10.53\scalebox{.6}{±0.27} & 44.19\scalebox{.6}{±0.79} & 42.13\scalebox{.6}{±0.27} & 31.12\scalebox{.6}{±0.75} & 29.63\scalebox{.6}{±0.31} & 43.20\scalebox{.6}{±1.24} & 39.61\scalebox{.6}{±0.14} & 26.66 & 3.3 & 3 \\

\hspace{3mm} --- $\mathbb{1}_{FreeMulti}$ & 7.23\scalebox{.6}{±0.18} & 7.02\scalebox{.6}{±0.24} & 18.41\scalebox{.6}{±5.86} & 11.28\scalebox{.6}{±0.86} & 44.58\scalebox{.6}{±0.96} & 42.08\scalebox{.6}{±0.29} & 36.15\scalebox{.6}{±1.36} & 30.35\scalebox{.6}{±0.70} & 43.94\scalebox{.6}{±1.68} & 39.16\scalebox{.6}{±0.36} & 28.02 & 4.6 & 5 \\

\hspace{3mm} --- $\mathbb{1}_{Multi}$ & \bluebold{6.88\scalebox{.6}{±0.14}} & 7.46\scalebox{.6}{±0.50} & 15.77\scalebox{.6}{±3.04} & 11.66\scalebox{.6}{±0.18} & 44.57\scalebox{.6}{±0.80} & 42.31\scalebox{.6}{±0.36} & 32.13\scalebox{.6}{±0.45} & 30.35\scalebox{.6}{±0.26} & 44.68\scalebox{.6}{±0.89} & 39.99\scalebox{.6}{±0.22} & 27.58 & 5.2 & 6 \\

\hspace{3mm} --- $\gamma_{min}$ & 7.09\scalebox{.6}{±0.24} & \bluebold{6.98\scalebox{.6}{±0.20}} & 11.18\scalebox{.6}{±0.66} & \bluebold{10.56\scalebox{.6}{±0.63}} & \bluebold{44.11\scalebox{.6}{±0.79}} & \bluebold{41.99\scalebox{.6}{±0.18}} & 31.14\scalebox{.6}{±0.61} & 29.59\scalebox{.6}{±0.27} & 43.37\scalebox{.6}{±0.65} & 39.17\scalebox{.6}{±0.31} & 26.52 & 2.4 & 2\\

\bottomrule
\end{tabular}
\caption{Ablation study by removing one component from MultiMatch. The best results are highlighted in \bluebold{blue}.}
\label{tab:ablation-results}
\vspace{-3mm}
\end{table*}
MultiMatch outperforms all the baselines, ranking first in 7 out of 10 setups and outperforming the second-best approach, FreeMatch, by 3.26\% averaged across all tasks. In particular, MultiMatch reaches state-of-the-art performance in all five setups with the imbalance factor 100, when the unlabeled distribution is similar to the labeled one, and surpasses FreeMatch by 3.48\% when the unlabeled distribution is different. Comparing to MarginMatch, the main competitor in the balanced setup, MultiMatch surpasses it by an average of 8.1\% in the imbalanced setting, with an impressive 11.74\% improvement when the unlabeled distribution differs from the labeled one. 

The superior performance of MultiMatch in highly imbalanced settings can be attributed to several CISSL techniques incorporated into its design.
First, the use of class-wise thresholds enables the algorithm to account for the distinct learning progress of individual classes. For the same reason, FreeMatch shows a substantial improvement in imbalanced setups, ranking second before the ABC enhancement.
Second, the use of percentile-based thresholds mitigates imbalance amplification, as shown by \citet{guo2022class}.
Finally, MultiMatch explicitly promotes lower pseudo-label impurity through several mechanisms. The use of multiple filters increases the difficulty for incorrect pseudo-labels to pass, while the subset employed for threshold computation results in generally higher thresholds compared to MarginMatch. 
This approach helps mitigate long-term error accumulation, a key challenge in imbalanced SSL. For a more detailed technical discussion, see Appendix~\ref{sec:appendix:multimatch_cissl_design}.


Enhancing SSL algorithms with ABC improves the overall performance of each algorithm. Although MarginMatch benefits the most with a 5.97\% improvement and MultiMatch the least with only 0.91\%, MultiMatch remains the best-performing approach, achieving state-of-the-art results on five setups and an error rate 1.8\% lower than FreeMatch+ABC, the second-best ABC-enhanced method. Notably, even without ABC enhancement, MultiMatch still outperforms all other ABC-enhanced algorithms in terms of mean error. This highlights the robustness of our approach to imbalanced class distributions, a trait that is essential for real-world text classification datasets.



\section{Ablation Study and Analysis}
\label{sec_ablation}

To demonstrate the effectiveness of each component in MultiMatch, we evaluate its performance after systematically removing different elements, as presented in Table~\ref{tab:ablation-results}.
We observe a decline in performance after the removal of each, indicating that all components in MultiMatch play a significant role in achieving optimal performance.

First, in order to assess the importance of \textit{useful \& difficult} examples, we remove the weight $w_d$, so they will be weighted similar to the \textit{useful \& easy} samples. This leads to a drop in performance on 9 out of 10 setups, with a mean error rate increase of 0.3\%.
Similarly, by completely removing these examples from the training data, on average we do not notice any substantial differences.
Second, we remove the filtering based on the self-adaptive thresholds that rely on the current confidence of the model, $\mathbb{1}_{FreeMulti}$. This results in our PLWM using only the heads agreement and the APM score for filtering pseudo-labels. This leads to a significant drop in performance of 1.65\%, highlighting the importance of current model's beliefs in our PLWM.
Third, we remove the APM-based filtering $\mathbb{1}_{Multi}$ from MultiMatch, obtaining a PLWM module that only uses the agreement samples where at least one head has high current confidence in the selected pseudo-label. In this case, the mean error rate increases by 1.21\%, and the Friedman test places it the lowest among all the setups in the ablation study.
Finally, removing the low limit for the class-wise thresholds 
($\gamma_{min}$)
has the lowest impact of only 0.15\% error rate, but it still provides additional proof that a stricter filtering mechanism is essential for reducing the noise in the pseudo-labels and preventing their degradation over time.

\begin{figure}
    \captionsetup{font=small}
    \centering
    \begin{subfigure}[b]{0.23\textwidth}
    \includegraphics[width=\textwidth]{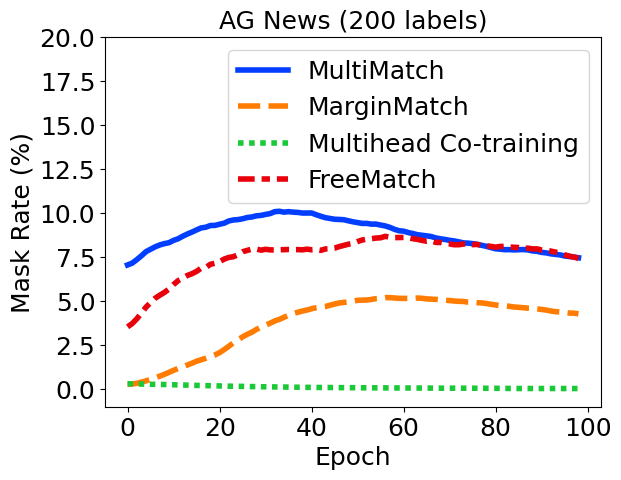}
  \end{subfigure} 
    \hfill
    \begin{subfigure}[b]{0.23\textwidth}
      \centering
    \includegraphics[width=\textwidth]{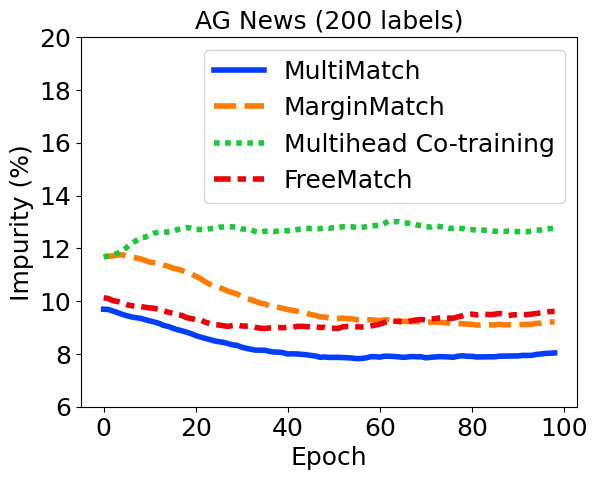} 
      \end{subfigure} 
      \caption{Mask rate and impurity on AG News with 200 labels.} 
    \label{fig:stats_agnews_balanced}
    \vspace{-3mm}
\end{figure}

\begin{table*}[!htbp]
\scriptsize
\centering 
\captionsetup{font=small}
\tabcolsep=0.15cm

\begin{tabular}{l|cccc|cccc}
\toprule
Task & \multicolumn{4}{c|}{Humanitarian} & \multicolumn{4}{c}{Informative} \\
 & {Acc} & {P} & {R} & {F1} & {Acc} & {P} & {R} & {F1} \\

\midrule

{MMBT Supervised}~\citep{kiela2019supervised} & 86.71 & 87.20 & 86.75 & 86.74 & 89.44 & 90.07 & 90.06 & 89.87\\
\midrule
{FixMatch}~\citep{sohn2020fixmatch} & 88.55 & 88.87 & 88.59 & 88.51 & 89.96 & 89.91 & 90.00 & 89.91\\
{FixMatch LS}~\citep{sirbu-etal-2022-multimodal} & 88.66 & 89.04 & 88.70 & 88.74 & 90.38 & 90.35 & 90.42 & 90.36\\
{FreeMatch}~\citep{wang2023freematch} & 84.88 & 86.93 & 84.93 & 85.54 & 90.35 & 90.52 & 90.39 & 90.43\\
{Multihead Co-training}~\citep{chen2021semisupervised} & 88.34 & 88.68 & 88.38 & 88.28 & 90.68 & 90.66 & 90.71 & 90.67\\
{MarginMatch}~\citep{sosea2023marginmatch} & 87.39 & 88.48 & 87.43 & 87.54 & 89.86 & 90.12 & 89.86 & 89.94\\

\midrule

\textbf{MultiMatch} & \bluebold{89.18} &	\bluebold{89.58} &	\bluebold{89.18}	& \bluebold{89.14} & \bluebold{91.36} &	\bluebold{91.37} &	\bluebold{91.36} &	\bluebold{91.37} \\

\bottomrule
\end{tabular}
\caption{\redold{Classification results on the multimodal \textit{Humanitarian} and \textit{Informative} CrisisMMD tasks. The best result for each metric is highlighted in \bluebold{blue}. \textbf{Acc} represents the accuracy, while \textbf{P}, \textbf{R} and \textbf{F1} are the weighted precision, recall and F1, respectively.}}
\vspace{-4mm}
\label{tab:crisismmd-results}
\end{table*}

\paragraph{Pseudo-Label Quality Analysis.} 
We compare MultiMatch with MarginMatch, Multihead Co-training, and FreeMatch in terms of pseudo-label quality using two metrics: \textit{mask rate} and \textit{impurity}. 
We show the results on the AG News task with 200 labels 
in Figure \ref{fig:stats_agnews_balanced} (similar results are shown in Figure \ref{fig:stats_agnews_imb100} in Appendix~\ref{sec:appendix:multimatch_cissl_design} for the imbalanced setting). 
The \textit{mask rate} is defined as the proportion of pseudo-labeled examples that were excluded from training at epoch $t$, while \textit{impurity} is the proportion of pseudo-labeled examples that participate in training at epoch $t$ but have an incorrect label.
As expected, our PLWM module employs a stricter filtering mechanism, resulting in a higher mask rate than all other approaches. In contrast, relying only on heads agreement proves ineffective, as Multihead Co-training exhibits a mask rate of less than 1\%. Meanwhile, our strict filtering, combined with a targeted approach for handling \textit{difficult} examples, effectively minimizes impurity in both setups. In contrast, Multihead Co-training incorporates many erroneous pseudo-labels because of its weak filtering, while the self-adaptive thresholds of FreeMatch fail to prevent error accumulation in the imbalanced setup, causing impurity to rise. 



\section{Performance in Multimodal Settings}
\label{sec:appendix:multimodal}

We evaluate the generalizability of MultiMatch in real-world multimodal settings by applying it to the multimodal version of the CrisisMMD dataset \cite{multimodalbaseline2020} and compare its performance against several representative baselines. We adopt the experimental setup from \citet{sirbu-etal-2022-multimodal}, which includes access to unlabeled data suitable for SSL. Specifically, we use MMBT~\cite{kiela2019supervised} as the supervised backbone and follow the same hyperparameter configuration as in their work. Importantly, we apply MultiMatch with the same hyperparameters used in the NLP experiments, without any task-specific tuning.

We selected CrisisMMD for several reasons. First, it provides real-world data from the crisis response domain. Second, the dataset is multimodal -- text and images -- unlike the previously used NLP datasets. Third, it supports multiple classification tasks with distinct characteristics: for example, the Informative task features a nearly balanced binary classification setup, while the Humanitarian task involves five classes and is highly imbalanced. Additional data details can be found in \citet{multimodalbaseline2020}.

As shown in Table~\ref{tab:crisismmd-results}, MultiMatch demonstrates strong generalization, consistently outperforming all other SSL baselines across both tasks. In contrast, other methods exhibit inconsistent performance. For instance, FreeMatch and Multihead Co-training perform well on the Informative task but struggle on the Humanitarian task. MarginMatch, which previously showed competitive results on NLP datasets, is outperformed by several alternatives in this multimodal setting.
While we do not delve into a deeper investigation of domain-specific challenges, the results highlight the robustness of MultiMatch in multimodal, domain-specific, and highly imbalanced setups.


\section{Conclusion}
\label{sec_conclusion}

In this work, we introduced MultiMatch, a novel semi-supervised learning algorithm that integrates co-training, consistency regularization, and pseudo-labeling into a unified framework.
Our approach effectively balances pseudo-label selection, filtering, and weighting, enhancing both robustness and performance. Extensive experiments on the USB benchmark demonstrate that MultiMatch consistently outperforms 18 baseline methods
in both balanced and highly imbalanced settings, which are crucial for real-world text classification tasks. Furthermore, our ablation study highlights the significance of each component in MultiMatch, demonstrating that our strict three-fold filtering mechanism is crucial for minimizing erroneous pseudo-labels and enhancing performance. These results emphasize the effectiveness of hybrid SSL strategies and open new directions for developing more adaptive and robust 
techniques.

\vspace{-1mm}
\section*{Acknowledgements}
\vspace{-1mm}
This research is supported by the project “Romanian Hub for Artificial Intelligence - HRIA”, Smart Growth, Digitization and Financial Instruments Program, 2021-2027, MySMIS no. 334906; the NSF IIS award 2107518; and a UIC Discovery Partners Institute (DPI) award. Any opinions, findings, and conclusions expressed here are those of the authors and do not necessarily reflect the views of NSF or DPI. We also thank our anonymous reviewers for their insightful feedback and suggestions.

\section*{Limitations}

While MultiMatch demonstrates strong performance across various setups, it has certain limitations. First, its reliance on multiple classification heads increases computational overhead compared to single-head SSL methods. In our case, the number of parameters increased by roughly 1\% (from 110M to 111M), but this may vary depending on the backbone. However, this overhead is negligible compared to traditional co-training approaches that train two distinct models. We extend the efficiency considerations in Appendix~\ref{sec:appendix:efficiency}.
Second, applying MultiMatch to specialized applications may require domain-specific knowledge integration (e.g., customized augmentation strategies) and additional hyperparameter tuning. However, this limitation is shared by all SSL methods. In Section~\ref{sec:appendix:multimodal}, we take initial steps toward assessing the generalizability of MultiMatch by applying it in a multimodal, real-world experimental setting. The results highlight the particular robustness of MultiMatch across two distinct tasks, even without any hyperparameter tuning, in contrast to other baseline methods. Nonetheless, further analysis remains advisable for highly sensitive applications.
\bibliography{main}

\appendix

\section{Class-imbalanced Semi-supervised Learning}
\label{sec:appendix:abc}

Existing semi-supervised learning algorithms usually assume a balanced class distribution of the data, which is not usually the case in real-world scenarios. Furthermore, while classifiers in general have a bias toward the majority classes, this poses a bigger problem for the SSL algorithms relying on pseudo-labeling. As the increasingly biased predictions are used for generating pseudo-labels, the bias increases even more.
To address this, ABC \cite{lee2021abc} proposes an auxiliary balanced classifier that can be attached to an existing SSL algorithm in order to mitigate class imbalance. The auxiliary classifier is attached to the last features layer of the model so it can leverage useful representations learned by the backbone from the entire dataset. However, it is trained by masking the samples with a probability inversly proportional to the size of their corresponding class, thus enabling the use of a balanced classification loss. Note that the ABC loss is just a supplementary loss over the base SSL algorithm. This makes the approach different from plain undersampling of the majority class, as it is still used in it's entirety by the base SSL algorithm.

Due to the significant performance improvement shown by ABC on class-imbalanced datasets from Computer Vision using FixMatch \cite{sohn2020fixmatch} as a base SSL algorithm and due to its versatility, as it can be used in combination with any base SSL algorithm, we also include it in our experiments. As our MultiMatch approach obtains particularly impressive results over the baselines in highly imbalanced setups, we find it relevant to assess whether the gap between our approach and other SSL algorithms can be bridged simply by the integration of a balancing method such as ABC in the training pipeline.

To this end, for each SSL algorithm tested in the highly imbalanced setups, we perform additional experiments using the same algorithm in combination with ABC \cite{lee2021abc} and show the results in Table \ref{tab:imbalanced-results}.

\section{MultiMatch for Class Imbalance}
\label{sec:appendix:multimatch_cissl_design}

\redold{According to \citet{gui2024survey_cissl}, CISSL methods address the bias accumulation problem in SSL algorithms through two main strategies. The first approach, already described in Appendix~\ref{sec:appendix:abc} and exemplified by ABC~\cite{lee2021abc}, adjusts the classifier to mimic a balanced class distribution. In this setup, all unlabeled samples contribute to learning useful representations, but the classification loss is computed in a way that accounts for class imbalance -- typically by weighting or masking examples based on class frequency.
The second, more direct strategy aims to improve pseudo-label quality by reducing the influence of overrepresented classes. For instance, CReST~\cite{wei2021crest} modifies the self-training paradigm with a re-sampling technique that prioritizes pseudo-labels from minority classes. Meanwhile, Adsh~\cite{guo2022class} introduces an adaptive thresholding mechanism that enforces the use of a similar percentage of pseudo-labels for each class, mitigating imbalance amplification.
Finally, reducing pseudo-label impurity -- defined as the labeling error at each epoch -- can help limit long-term error accumulation.}

\redold{In this context, MultiMatch incorporates, by design, several CISSL techniques that explain its superior performance in highly imbalanced settings. First, the use of class-wise thresholds allows the algorithm to differentiate between the learning progress of individual classes. The impact of this thresholding strategy is evident in Table~\ref{tab:imbalanced-results}, by analyzing the performance of FreeMatch. While it performs similarly to FixMatch under balanced conditions, it shows a substantial improvement in imbalanced setups, reaching second place before applying the ABC enhancement.}

\redold{Second, MultiMatch uses a percentile-based threshold for each class, similar to Adsh, enabling a more equitable selection of pseudo-labels across classes. Although MarginMatch also employs a percentile-based threshold, it fails to achieve the same effect due to its lack of class-wise computation.}

\begin{figure}
\vspace{-1mm}
    \captionsetup{font=small}
    \centering
    \begin{subfigure}[b]{0.23\textwidth}
    \includegraphics[width=\textwidth]{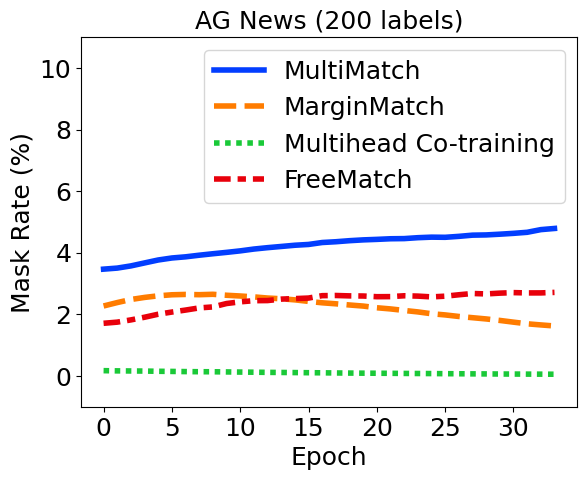}
  \end{subfigure} 
    \hfill
    \begin{subfigure}[b]{0.23\textwidth}
      \centering
    \includegraphics[width=\textwidth]{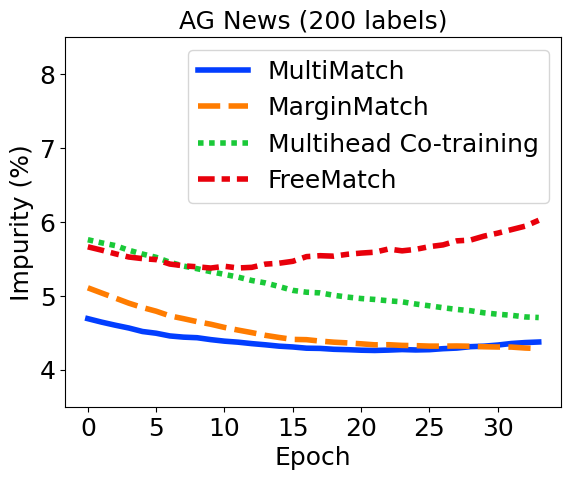} 
      \end{subfigure} 
      \caption{Mask rate (left) and impurity (right) on AG News with imbalance factor 100.} 
    \label{fig:stats_agnews_imb100}
    \vspace{-2mm}
\end{figure}

\redold{Finally, MultiMatch explicitly promotes lower pseudo-label impurity through several mechanisms. The use of multiple filters increases the difficulty for incorrect pseudo-labels to pass, while the subset employed for threshold computation results in generally higher thresholds compared to MarginMatch. 
As shown in Figure~\ref{fig:stats_agnews_imb100}, MultiMatch maintains a low pseudo-label impurity -- therefore mitigating error accumulation -- using a higher mask rate of the predictions. While \citet{sosea2023marginmatch} point out that excessive masking can discard informative borderline examples, MultiMatch compensates for this by assigning a higher weight to the \textit{useful \& difficult} samples that pass all filtering criteria.}

\section{Extended Background}
\label{sec:extended_background}

FreeMatch \cite{wang2023freematch} shows that using a fixed confidence threshold $\tau$ is suboptimal, as it does not account for varying difficulty across classes and for the model's increase in confidence over time. They address these issues by introducing class-wise thresholds, adaptively scaling $\tau(c)$ depending on the learning status of each class. Specifically, FreeMatch computes a self-adaptive global threshold $\tau_t$ that reflects the confidence
of the model on the whole unlabeled dataset:

\begin{equation}
    \tau_t = \lambda \tau_{t-1} + (1 - \lambda) \frac{1}{\mu B} \sum_{b=1}^{\mu B} \max(q_b)
    \label{eq_freematch_global_thr}
\end{equation}

\noindent
and a self-adaptive local threshold $\tilde{p}_t(c)$ that reflects the confidence of the model for a specific class:
\begin{equation}
    \tilde{p}_t(c) = \lambda \tilde{p}_{t-1}(c) + (1 - \lambda) \frac{1}{\mu B} \sum_{b=1}^{\mu B} q_b(c)
    \label{eq_freematch_local_thr}
\end{equation}
\noindent
with $\tau_0 = \tilde{p}_0(c) = \frac{1}{C}$.

Finally, the two thresholds are combined in order to obtain the final self-adaptive class thresholds $\tau_t(c)$ as the product between the normalized local thresholds and the global one.

\begin{equation}
    \tau_t(c) = \frac{\tilde{p_t}(c)}{\displaystyle \max_{c'}(\tilde{p_t}(c'))} \cdot \tau_t
    \label{eq_freematch_final_thr}
\end{equation}

\noindent
These thresholds are used to replace the filtering function from FixMatch with:

\begin{equation}
    \mathbb{1}_{Free}^{(t)} = \mathbb{1}(\max(q_b) > \tau_t(\hat{q_b}))
    \label{eq_freematch_filter_appendix}
\end{equation}

\section{Experimental Setup}
\label{sec:hyperparams}

\paragraph{Shared Hyperparameter Settings.}
In order to ensure a straightforward and fair comparison with the algorithms already included in USB \cite{wang2022usb}, we follow their experimental setup and use the same hyperparameters in all the experiments. More precisely, we use the pretrained uncased version of BERT-Base \cite{devlin2018bert} with a maximum text length of 512 tokens for all the algorithms. While suboptimal, we use no weak augmentation ($\alpha(x) = x$), to ensure backward compatibility. For strong augmentation we adopt back-translation \cite{xie2020unsupervised_bt} with De-En and Ru-En translation using WMT19. While for the other 4 datasets the augmentations were already provided in USB so they were identical in all experiments, for IMDB we had to regenerate the translations, so there might be differences between the translations used by the 16 baselines from USB and the 3 methods implemented by us into the framework, due to the temperature 0.9 used for translation. 
The optimization with the AdamW optimizer with weight decay of 1e-4, cosine learning rate scheduler with a total training steps of 102, 400 and a warm-up of 5, 120 steps were preserved in all experiments. The optimal learning rates and layer decay rates for each dataset were determined by \citet{wang2022usb} using a grid search for the FixMatch \cite{sohn2020fixmatch} approach and were used unchanged for all approaches. The only change in setup compared to those reported in USB \cite{wang2022usb} was the use of batch size 8 instead of 4, as it was shown to improve performance \footnote{\scriptsize{\url{https://github.com/microsoft/Semi-supervised-learning/blob/main/results/usb_nlp.csv}}}. The reported results for the baselines also used this improved batch size. Following USB, we also used the unlabeled loss weight $w_u=1$ and the ration between unlabeled and labeled examples in a batch $\mu=1$, although it was shown that a larger $\mu$ improves performance \cite{sohn2020fixmatch, sirbu-etal-2022-multimodal}.

\paragraph{MultiMatch Hyperparameters.}

The hyperparameters specific to MultiMatch are the number of heads $H$, which we set to 3 from the start, similar to Multihead Co-training, the weight $w_d$ used for the \textit{useful \& difficult} samples, the percentile $f$ used in computing our self-adaptive APM thresholds, and their lower bound $\gamma_{min}$. For the APM percentile $f$ we experimented with values of 5\% and 10\%, and selected 5\% as the optimal value. Setting $f$=10\% determined an increased mask rate, which excluded relevant samples from the unsupervised loss computation. For the lower bound $\gamma_{min}$, we experimented with the values 0 and $-\infty$ (no lower bound), and found that $-\infty$ works better in highly imbalanced setups as it allows for more pseudo-labels from the underrepresented classes. For the weight $w_d$ of the \textit{useful \& difficult} examples, we experimented with values of 3 and 5, selecting 3 as the optimal value. The hyperparameters selection was done following experiments on the AG News dataset with 200 labels and choosing the value that resulted in a smaller validation error averaged over 3 runs. The method was chosen due to the faster training time on AG News, but it does not exclude that different values might be optimal for other tasks. However, our selected values provided consistent performance across all setups.
\begin{table*}[htbp]
\scriptsize
\centering 
\tabcolsep=0.15cm
\captionsetup{font=small}

\begin{tabular}{c|c|c|c|c|c|c}
\toprule

Dataset & Label Type & \#Classes & \#Labeled & \#Unlabeled  & \#Validation & \#Test \\

\midrule

IMDB & Movie Review Sentiment & 2 & 10 / 50 & 11,500 & 1,000 & 12,500 \\
AG News & News Topic & 4 & 10 / 50 & 25,000 & 2,500 & \hphantom{0}1,900 \\
Amazon Review & Product Review Sentiment & 5 & 50 / 200 & 50,000 & 5,000 & 13,000 \\
Yahoo! Answer & QA Topic & 10 & 50 / 200 & 50,000 & 5,000 & \hphantom{0}6,000 \\
Yelp Review &  Restaurant Review Sentiment & 5 & 50 / 200 & 50,000 & 5,000 & 10,000 \\

\bottomrule
\end{tabular}
\caption{Dataset statistics and split information as number of examples per class.}
\label{tab:dataset-statistics}
\end{table*}
\paragraph{Computational Resources.}
We conduct our experiments on a system with 8 Nvidia H100 GPUs with 80GB VRAM. Each experiment was run in a single GPU setting, the multiple GPUs being used to run multiple experiments simultaneously. 
Each training run took between 2 and 4 hours depending on the 
dataset ($\approx$2h for AG News, $\approx$3h for IMDB and Yahoo, and $\approx$4h for Amazon and Yelp). With an average of $\approx$3h per run, 3 runs per setup, 10 setups (5 datasets, 2 sizes each) and 5 models (MultiMatch, MarginMatch, Multihead Co-training, SequenceMatch, CGMatch), the total training time for the main results table is about 3x3x10x5 = 450 GPU hours. For the imbalanced setups we trained 10 models (5 models with and without ABC) for a total of 3x3x10x10 = 900 GPU hours). The ablation study required to train 5 additional models for a total of 3x3x10x5 = 450 GPU hours.
Adding this up, the total cost was about 1,800 GPU hours, or 75 GPU days.

\section{Datasets}
\label{sec:appendix_dataset}

Statistics for the text classification datasets employed in this work are shown in Table~\ref{tab:dataset-statistics}.

For the highly imbalanced setups, we use between 1,000 and 10 labels per class according to a long-tail distribution, as described in Section~\ref{sec:experiments}. For the imbalance factor 100 we use 10 times more unlabeled data for each class compared to the labeled data. For the imbalance factor -100, the total number of unlabeled samples is still 10 times higher than the labeled one, but the long-tail distribution of the unlabeled set is reversed. The validation and test sets are the same as the balanced settings.

\section{Efficiency Considerations}
\label{sec:appendix:efficiency}

\redold{While SSL algorithms may significantly boost the performance of a supervised model by leveraging additional unlabeled data, they typically lead to slower training, as more data must be processed. Specifically, for a supervised model, the total training time $\mathcal{T}$ is directly proportional to the number of epochs $N_e$ and the number of labeled examples $N_L$ (i.e., $\mathcal{T} \approx N_L \times N_e$). However, in the case of an SSL approach, since each epoch processes $(\mu+1)\times N_L$ examples, the total training time becomes $\mathcal{T} \approx (\mu+1)\times N_L \times N_e$. This applies to all SSL algorithms used in this work.}

\redold{Furthermore, some design choices may introduce additional computational overhead. For instance, a traditional co-training method requires twice the time and memory, as it trains two models in parallel. While Multihead Co-Training addresses this by sharing a backbone across heads, the presence of three classification heads still implies extra computation through two additional output layers. However, classification heads are typically small, and the impact on runtime varies depending on the backbone. In our experiments using BERT as the shared backbone, the training overhead of Multihead Co-Training and MultiMatch remained consistently below $1\%$ compared to FixMatch.}

\redold{At inference time, models trained with SSL behave similarly to their supervised counterparts, as the same amount of data is passed through the model. However, the overhead of a multihead architecture still persists. In our setup, predictions are made by averaging the logits from all three heads, which leads to a minor inference overhead (approximately $1\%$) compared to single-head architectures. While this may be relevant for real-world deployments, the overhead can be removed after training by retaining only a single head during inference. This incurs a marginal performance drop of approximately $0.1\%$ on our tasks, as the three heads converge to high agreement in the final training stages.}

\section{Results}
\label{sec:appendix}

Due to space constraints, we removed the 7 weakest baselines from Table~\ref{tab:full-results}.
The full version of Table \ref{tab:full-results}, which contains all baselines, is provided in Table \ref{tab:full_full-results}.
A summary of the results, together with the citations for each SSL algorithm, is provided in Table \ref{tab:citations}. For the \redold{shared} backbone, we used the uncased BERT \cite{devlin2018bert} in all experiments.

\redold{MultiMatch achieves statistically significant improvements over USB baselines, as confirmed by a Friedman test followed by the Nemenyi post-hoc test \cite{demvsar2006statistical}, which is a standard methodology for comparing multiple classifiers across multiple datasets.
The Friedman test first evaluates whether there are significant differences among the classifiers' rankings across datasets. Upon finding a statistically significant difference, the Nemenyi test is then applied to identify which pairs of classifiers differ significantly. Our results indicate that MultiMatch performs significantly better than USB baselines, highlighting the robustness and generalizability of our proposed approach across a diverse range of text classification tasks.}

\begin{table*}[!htbp]
\scriptsize
\centering 
\captionsetup{font=small}
\tabcolsep=0.1cm

\begin{tabular}{l|rr|rr|rr|rr|rr|r|c|c}
\toprule

Dataset & \multicolumn{2}{c|}{IMDB} & \multicolumn{2}{c|}{AG News} & \multicolumn{2}{c|}{Amazon Review} & \multicolumn{2}{c|}{Yahoo! Answer} & \multicolumn{2}{c|}{Yelp Review}  & Mean & Fried. & Final \\

\# Label & 20 & 100 & 40 & 200 & 250 & 1000 & 500 & 2000 & 250 & 1000 & error & rank & rank \\

\midrule

Supervised (Full) & 5.69\scalebox{.6}{±0.15} & 5.72\scalebox{.6}{±0.13} & 5.78\scalebox{.6}{±0.07} & 5.73\scalebox{.6}{±0.11} & 36.40\scalebox{.6}{±0.05} & 36.40\scalebox{.6}{±0.05} & 24.87\scalebox{.6}{±0.04} & 24.84\scalebox{.6}{±0.04} & 32.04\scalebox{.6}{±0.03} & 32.04\scalebox{.6}{±0.03} & 20.95 & - & -\\
Supervised (Small) & 20.31\scalebox{.6}{±2.79} & 14.02\scalebox{.6}{±1.22} & 15.06\scalebox{.6}{±1.08} & 14.25\scalebox{.6}{±0.97} & 52.31\scalebox{.6}{±1.28} & 47.53\scalebox{.6}{±0.69} & 37.43\scalebox{.6}{±0.29} & 33.26\scalebox{.6}{±0.10} & 51.22\scalebox{.6}{±0.98} & 46.71\scalebox{.6}{±0.37} & 33.21 & - & -\\
\midrule
Pseudo-Labeling & 45.45\scalebox{.6}{±4.43} & 19.67\scalebox{.6}{±1.01} & 19.49\scalebox{.6}{±3.07} & 14.69\scalebox{.6}{±1.88} & 53.45\scalebox{.6}{±1.90} & 47.00\scalebox{.6}{±0.79} & 37.70\scalebox{.6}{±0.65} & 32.72\scalebox{.6}{±0.31} & 54.51\scalebox{.6}{±0.82} & 47.33\scalebox{.6}{±0.20} & 37.20 & 17.3 & 18 \\
Mean Teacher  & 20.06\scalebox{.6}{±2.51} & 13.97\scalebox{.6}{±1.49} & 15.17\scalebox{.6}{±1.21} & 13.93\scalebox{.6}{±0.65} & 52.14\scalebox{.6}{±0.52} & 47.66\scalebox{.6}{±0.84} & 37.09\scalebox{.6}{±0.18} & 33.43\scalebox{.6}{±0.28} & 50.60\scalebox{.6}{±0.62} & 47.21\scalebox{.6}{±0.31} & 33.13 & 16.0 & 16 \\
$\Pi$-Model & 49.99\scalebox{.6}{±0.01} & 44.75\scalebox{.6}{±3.99} & 60.70\scalebox{.6}{±19.0} & 12.58\scalebox{.6}{±0.57} & 77.22\scalebox{.6}{±1.50} & 53.17\scalebox{.6}{±2.56} & 44.91\scalebox{.6}{±1.32} & 32.45\scalebox{.6}{±0.45} & 75.73\scalebox{.6}{±4.01} & 59.82\scalebox{.6}{±0.61} & 51.13 & 18.6 & 19 \\
VAT & 25.93\scalebox{.6}{±2.58} & 11.61\scalebox{.6}{±1.79} & 14.70\scalebox{.6}{±1.19} & 11.71\scalebox{.6}{±0.84} & 49.83\scalebox{.6}{±0.46} & 46.54\scalebox{.6}{±0.31} & 34.87\scalebox{.6}{±0.41} & 31.50\scalebox{.6}{±0.35} & 52.97\scalebox{.6}{±1.41} & 45.30\scalebox{.6}{±0.32} & 32.50 & 13.6 & 15 \\
MixMatch & 26.12\scalebox{.6}{±6.13} & 15.47\scalebox{.6}{±0.65} & 13.50\scalebox{.6}{±1.51} & 11.75\scalebox{.6}{±0.60} & 59.54\scalebox{.6}{±0.67} & 61.69\scalebox{.6}{±3.32} & 35.75\scalebox{.6}{±0.71} & 33.62\scalebox{.6}{±0.14} & 53.98\scalebox{.6}{±0.59} & 51.70\scalebox{.6}{±0.68} & 36.31 & 16.1 & 17 \\
ReMixMatch & 50.00\scalebox{.6}{±0.00} & 50.00\scalebox{.6}{±0.00} & 75.00\scalebox{.6}{±0.00} & 75.00\scalebox{.6}{±0.00} & 80.00\scalebox{.6}{±0.00} & 80.00\scalebox{.6}{±0.00} & 90.00\scalebox{.6}{±0.00} & 90.00\scalebox{.6}{±0.00} & 80.00\scalebox{.6}{±0.00} & 80.00\scalebox{.6}{±0.00} & 75.00 & 20.9 & 21 \\
AdaMatch & 8.09\scalebox{.6}{±0.99} & 7.11\scalebox{.6}{±0.20} & 11.73\scalebox{.6}{±0.17} & 11.22\scalebox{.6}{±0.95} & 46.72\scalebox{.6}{±0.72} & 42.27\scalebox{.6}{±0.25} & 32.75\scalebox{.6}{±0.35} & 30.44\scalebox{.6}{±0.31} & 45.40\scalebox{.6}{±0.96} & 40.16\scalebox{.6}{±0.49} & 27.59 & \hphantom{0}6.5 & \hphantom{0}5 \\
UDA & 49.97\scalebox{.6}{±0.04} & 50.00\scalebox{.6}{±0.00} & 41.00\scalebox{.6}{±24.9} & 53.68\scalebox{.6}{±30.1} & 60.76\scalebox{.6}{±13.6} & 68.38\scalebox{.6}{±16.4} & 71.30\scalebox{.6}{±26.4} & 70.50\scalebox{.6}{±27.5} & 69.33\scalebox{.6}{±15.0} & 66.95\scalebox{.6}{±18.4} & 60.19 & 19.6 & 20 \\
FixMatch & 7.72\scalebox{.6}{±0.33} & 7.33\scalebox{.6}{±0.13} & 30.17\scalebox{.6}{±1.87} & 11.71\scalebox{.6}{±1.95} & 47.61\scalebox{.6}{±0.83} & 43.05\scalebox{.6}{±0.54} & 33.03\scalebox{.6}{±0.49} & 30.51\scalebox{.6}{±0.53} & 46.52\scalebox{.6}{±0.94} & 40.65\scalebox{.6}{±0.46} & 29.83 & 10.2 & 11 \\
FlexMatch & 7.82\scalebox{.6}{±0.77} & 7.41\scalebox{.6}{±0.38} & 16.38\scalebox{.6}{±3.94} & 12.08\scalebox{.6}{±0.73} & 45.73\scalebox{.6}{±1.60} & 42.25\scalebox{.6}{±0.33} & 35.61\scalebox{.6}{±1.08} & 31.13\scalebox{.6}{±0.18} & 43.35\scalebox{.6}{±0.69} & 40.51\scalebox{.6}{±0.34} & 28.23 & \hphantom{0}9.4 & \hphantom{0}9 \\
Dash & 8.34\scalebox{.6}{±0.86} & 7.55\scalebox{.6}{±0.35} & 31.67\scalebox{.6}{±13.1} & 13.76\scalebox{.6}{±1.67} & 47.10\scalebox{.6}{±0.74} & 43.09\scalebox{.6}{±0.60} & 35.26\scalebox{.6}{±0.33} & 31.19\scalebox{.6}{±0.29} & 45.24\scalebox{.6}{±2.02} & 40.14\scalebox{.6}{±0.79} & 30.33 & 11.7 & 13 \\
CRMatch & 8.96\scalebox{.6}{±0.88} & 7.16\scalebox{.6}{±0.09} & 12.28\scalebox{.6}{±1.43} & 11.08\scalebox{.6}{±1.24} & 45.49\scalebox{.6}{±0.98} & 43.07\scalebox{.6}{±0.50} & 32.51\scalebox{.6}{±0.40} & 29.98\scalebox{.6}{±0.07} & 45.71\scalebox{.6}{±0.63} & 40.62\scalebox{.6}{±0.28} & 27.69 & \hphantom{0}7.6 & \hphantom{0}6 \\
CoMatch & 7.44\scalebox{.6}{±0.30} & 7.72\scalebox{.6}{±1.14} & 11.95\scalebox{.6}{±0.76} & 10.75\scalebox{.6}{±0.35} & 48.76\scalebox{.6}{±0.90} & 43.36\scalebox{.6}{±0.21} & 33.48\scalebox{.6}{±0.51} & 30.25\scalebox{.6}{±0.35} & 45.40\scalebox{.6}{±1.12} & 40.27\scalebox{.6}{±0.51} & 27.94 & \hphantom{0}8.1 & \hphantom{0}8 \\
SimMatch & 7.93\scalebox{.6}{±0.55} & 7.08\scalebox{.6}{±0.33} & 14.26\scalebox{.6}{±1.51} & 12.45\scalebox{.6}{±1.37} & 45.91\scalebox{.6}{±0.95} & 42.21\scalebox{.6}{±0.30} & 33.06\scalebox{.6}{±0.20} & 30.16\scalebox{.6}{±0.21} & 46.12\scalebox{.6}{±0.48} & 40.26\scalebox{.6}{±0.62} & 27.94 & \hphantom{0}7.6 & \hphantom{0}6 \\
FreeMatch & 8.94\scalebox{.6}{±0.21} & 7.95\scalebox{.6}{±0.45} & 12.98\scalebox{.6}{±0.58} & 11.73\scalebox{.6}{±0.63} & 46.41\scalebox{.6}{±0.60} & 42.64\scalebox{.6}{±0.06} & 32.77\scalebox{.6}{±0.26} & 30.32\scalebox{.6}{±0.18} & 47.95\scalebox{.6}{±1.45} & 40.37\scalebox{.6}{±1.00} & 28.21 & \hphantom{0}9.7 & 10 \\
SoftMatch & 7.76\scalebox{.6}{±0.58} & 7.97\scalebox{.6}{±0.72} & 11.90\scalebox{.6}{±0.27} & 11.72\scalebox{.6}{±1.58} & 45.29\scalebox{.6}{±0.95} & 42.21\scalebox{.6}{±0.20} & 33.07\scalebox{.6}{±0.31} & 30.44\scalebox{.6}{±0.62} & 44.09\scalebox{.6}{±0.50} & 39.76\scalebox{.6}{±0.13} & 27.42 & \hphantom{0}6.4 & \hphantom{0}4 \\
\midrule
Multihead Co-training & 8.70\scalebox{.6}{±0.88} & 7.46\scalebox{.6}{±0.68} & 22.72\scalebox{.6}{±6.09} & 13.48\scalebox{.6}{±1.51} & 46.22\scalebox{.6}{±0.70} & 43.07\scalebox{.6}{±0.79} & 35.17\scalebox{.6}{±1.58} & 30.81\scalebox{.6}{±0.28} & 46.46\scalebox{.6}{±1.28} & 40.79\scalebox{.6}{±0.49} & 29.49 & 12.2 & 14 \\
MarginMatch & 7.19\scalebox{.6}{±0.39} & 6.99\scalebox{.6}{±0.19} & \bluebold{10.65\scalebox{.6}{±0.19}} & 11.03\scalebox{.6}{±0.99} & 44.81\scalebox{.6}{±1.23} & 42.14\scalebox{.6}{±0.67} & 32.08\scalebox{.6}{±0.70} & 29.55\scalebox{.6}{±0.15} & 42.93\scalebox{.6}{±1.48} & 39.13\scalebox{.6}{±0.34} & 26.65 & \hphantom{0}2.3 & \hphantom{0}2 \\

SequenceMatch & 12.99\scalebox{.6}{±7.64} & 8.34\scalebox{.6}{±0.57} & 16.17\scalebox{.6}{±2.53} & 12.56\scalebox{.6}{±0.87} & 47.97\scalebox{.6}{±2.39} & 42.58\scalebox{.6}{±0.21} & 34.92\scalebox{.6}{±0.78} & 30.28\scalebox{.6}{±0.28} & 45.60\scalebox{.6}{±0.84} & 40.44\scalebox{.6}{±0.51} & 29.19 & 11.3 & 12 \\
CGMatch & 7.07\scalebox{.6}{±0.36} & \bluebold{6.79\scalebox{.6}{±0.10}} & 11.95\scalebox{.6}{±0.63} & 11.29\scalebox{.6}{±0.47} & 44.77\scalebox{.6}{±1.39} & 42.61\scalebox{.6}{±0.47} & 32.15\scalebox{.6}{±0.58} & 29.85\scalebox{.6}{±0.16} & 44.34\scalebox{.6}{±0.26} & 40.14\scalebox{.6}{±0.07} & 27.10 & \hphantom{0}3.9 & \hphantom{0}3 \\

\midrule
MultiMatch & \bluebold{6.89\scalebox{.6}{±0.07}} & 6.98\scalebox{.6}{±0.27} & 11.14\scalebox{.6}{±0.96} & \bluebold{10.59\scalebox{.6}{±0.66}} & \bluebold{44.43\scalebox{.6}{±0.98}} & \bluebold{42.09\scalebox{.6}{±0.28}} & \bluebold{30.90\scalebox{.6}{±0.70}} & \bluebold{29.39\scalebox{.6}{±0.39}} & \bluebold{42.16\scalebox{.6}{±0.79}} & \bluebold{39.08\scalebox{.6}{±0.55}} & \bluebold{26.37} & \hphantom{0}\bluebold{1.2} & \hphantom{0}\bluebold{1} \\

\bottomrule
\end{tabular}
\caption{Test error rates on IMDB, AG News, Amazon Review, Yahoo! Answer, and Yelp Review datasets using two setups with different sizes for the labeled set. The best result for each setup is highlighted in \bluebold{blue}. The upper section of the table contains results reported by the USB benchmark, while the bottom section contains baselines integrated by us into the USB codebase.}
\label{tab:full_full-results}
\end{table*}

\begin{table*}[htbp]
\scriptsize
\centering 
\captionsetup{font=small}

\begin{tabular}{c|c|c|c|c}
\toprule

Algorithm name & Citation  & Mean error & Friedman rank & Final rank \\


\midrule

BERT - Supervised (Full) & \citet{devlin2018bert} & 20.95 & - & -\\
BERT - Supervised (Small) & \citet{devlin2018bert} & 33.21 & - & -\\
\midrule
Pseudo-Labeling & \citet{lee2013pseudo_baseline_pseudolabeling} & 37.20 & 17.3 & 18 \\
Mean Teacher  & \citet{tarvainen2017mean_baseline_meanteacher} & 33.13 & 16.0 & 16 \\
$\Pi$-Model & \citet{rasmus2015semi_baseline_pimodel} & 51.13 & 18.6 & 19 \\
VAT & \citet{miyato2018virtual_baseline_vat} & 32.50 & 13.6 & 15 \\
MixMatch & \citet{berthelot2019mixmatch} & 36.31 & 16.1 & 17 \\
ReMixMatch & \citet{berthelot2019remixmatch} & 75.00 & 20.9 & 21 \\
AdaMatch & \citet{berthelot2021adamatch} & 27.59 & \hphantom{0}6.5 & \hphantom{0}5 \\
UDA & \citet{xie2020unsupervised_UDA} & 60.19 & 19.6 & 20 \\
FixMatch & \citet{sohn2020fixmatch} & 29.83 & 10.2 & 11 \\
FlexMatch & \citet{zhang2022flexmatch} & 28.23 & \hphantom{0}9.4 & \hphantom{0}9 \\
Dash & \citet{xu2021dash} & 30.33 & 11.7 & 13 \\
CRMatch & \citet{fan2023revisiting_baseline_crmatch} & 27.69 & \hphantom{0}7.6 & \hphantom{0}6 \\
CoMatch &\citet{li2021comatch} & 27.94 & \hphantom{0}8.1 & \hphantom{0}8 \\
SimMatch & \citet{zheng2022simmatch} & 27.94 & \hphantom{0}7.6 & \hphantom{0}6 \\
FreeMatch & \citet{wang2023freematch} & 28.21 & \hphantom{0}9.7 & 10 \\
SoftMatch & \citet{chen2023softmatch} & 27.42 & \hphantom{0}6.4 & \hphantom{0}4 \\
\midrule
Multihead Co-training & \citet{chen2021semisupervised} & 29.49 & 12.2 & 14 \\
MarginMatch & \citet{sosea2023marginmatch} & 26.65 & \hphantom{0}2.3 & \hphantom{0}2 \\

SequenceMatch & \citet{nguyen2024sequencematch} & 29.19 & 11.3 & 12 \\
CGMatch & \citet{cheng2025cgmatch} & 27.10 & \hphantom{0}3.9 & \hphantom{0}3 \\

\midrule
MultiMatch & Ours & \bluebold{26.37} & \hphantom{0}\bluebold{1.2} & \hphantom{0}\bluebold{1} \\

\bottomrule
\end{tabular}
\caption{Citations for all the methods used in our experiments as strong baselines.}
\label{tab:citations}
\end{table*}

\section{Comparison to Large Language Models}
\label{sec:appendix:LLMs_comparison}

Although a direct comparison between specialized SSL methods and general-purpose LLMs is inherently challenging (e.g., due to potential data contamination and significant disparity in model sizes), we include this analysis to provide a broader context for our proposed approach. These experiments are particularly valuable for assessing the performance of LLMs in specialized classification tasks and for highlighting the continued relevance of specialized, inference-efficient models in a landscape dominated by LLMs. To this end, we evaluate a representative LLM, GPT-4o-mini, on the five NLP datasets used in this work and show the results in Table~\ref{tab:gpt4o_results}.

\begin{table}[!htbp]
\scriptsize
\centering 
\captionsetup{font=small}
\begin{tabular}{l|c}
\toprule
Dataset & Test Error (\%) \\
\midrule
IMDB            & \hphantom{0}5.34 \\
AG News         & 16.26 \\
Amazon Review   & 36.42 \\
Yahoo! Answer   & 30.96 \\
Yelp Review     & 32.96 \\
\bottomrule
\end{tabular}
\caption{Performance of GPT-4o-mini on the USB datasets.}
\label{tab:gpt4o_results}
\end{table}

For the three sentiment classification tasks (IMDB, Amazon Review, and Yelp Review), the performance of GPT-4o-mini is comparable to the Supervised (Full) baseline presented in Table~\ref{tab:full-results}, which was trained with large quantities of labeled data (e.g. 250k samples for Amazon and Yelp). This strongly supports the hypothesis that such widely available datasets have been included in LLM training data, making them unsuitable for a fair comparison between SSL and LLMs.

Notably, MultiMatch significantly outperforms GPT-4o-mini on the AG News dataset. To understand the reasons behind this, we perform a quantitative and qualitative analysis of the LLM's predictions. The confusion matrix shown in Table~\ref{tab:gpt4o_confusion_matrix} reveals that the model frequently misclassifies news from the \textit{Sci/Tech} category as \textit{Business}, and occasionally \textit{World} news as \textit{Business}. A qualitative analysis shows that a large portion of the \textit{Sci/Tech} news involves corporate activities (e.g., acquisitions, mergers, layoffs, new project announcements). This semantic overlap with the \textit{Business} category leads to misclassification by the general-purpose LLM. Similarly, some \textit{World} news regarding inflation rates or trade agreements between governments are also mistakenly classified as \textit{Business}. These findings highlight the need for specialized training to distinguish between such nuanced categories, where SSL methods like MultiMatch prove effective even under limited labeled data.

\begin{table}[!htbp]

\scriptsize
\centering 
\captionsetup{font=small}
\begin{tabular}{l | c c c c}
\toprule
True $\backslash$ Pred & World & Sports & Business & Sci/Tech \\
\midrule
World     & 0.87 & 0.03 & 0.09 & 0.02 \\
Sports    & 0.01 & 0.98 & 0.01 & 0.00 \\
Business  & 0.03 & 0.00 & 0.92 & 0.04 \\
Sci/Tech  & 0.07 & 0.01 & 0.34 & 0.58 \\
\bottomrule
\end{tabular}
\caption{Confusion matrix of GPT-4o-mini predictions on AG News.}
\label{tab:gpt4o_confusion_matrix}
\end{table}

While these experiments provide valuable insights, we re-emphasize that a direct comparison between our SSL framework and a general-purpose LLM is inherently flawed. First, such a comparison needs to be performed in-depth, using novel datasets not seen during LLM pretraining. Secondly, our framework demonstrates the effectiveness of SSL for training small specialized models using limited labeled data. Such models have a significantly lower inference complexity than LLMs of billions of parameters, making them highly suitable for resource-constrained systems. As detailed in Appendix~\ref{sec:appendix:efficiency}, a model trained with SSL preserves the inference complexity of the underlying backbone (e.g., BERT). Therefore, the SSL domain remains highly relevant, providing a complementary approach to LLMs by offering high-performance solutions for specialized, resource-limited tasks.

\section{Code and License}

The source code for our MultiMatch method and all experimental configurations are publicly available on GitHub at \href{https://github.com/iustinsirbu13/MultiMatch}{https://github.com/iustinsirbu13/MultiMatch}. Our implementation is a public fork of the Unified Semi-supervised Learning (USB) framework \cite{wang2022usb} and is released under the MIT license, consistent with the original framework.
We have submitted a pull request to the official USB repository to integrate MultiMatch as a new algorithm, thereby contributing our work to the open-source community.





\end{document}